\documentclass[10pt,twocolumn,letterpaper]{article}

\usepackage{cvpr}              
\definecolor{cvprblue}{rgb}{0.21,0.49,0.74}
\usepackage[pagebackref,breaklinks,colorlinks,allcolors=cvprblue]{hyperref}

\usepackage{adjustbox}
\usepackage{multirow}
\usepackage{makecell}    
\usepackage{pifont}      
\usepackage{comment} 
\usepackage{tcolorbox}
\usepackage[skip=4pt]{caption} 

\newcommand{\cmark}{\ding{51}}%
\newcommand{\xmark}{\ding{55}}%

\newcommand{\myheading}[1]{%
  \vspace{0.5em}
  \noindent\textbf{#1.}\hspace{0.5em}%
}

\usepackage[accsupp]{axessibility}

\def\paperID{43300} 
\def\confName{CVPR}
\def\confYear{2026}

\title{PolySLGen: Online Multimodal Speaking-Listening Reaction Generation in Polyadic Interaction}

\author{Zhi-Yi Lin\textsuperscript{1}\quad Thomas Markhorst\textsuperscript{1} \quad Jouh Yeong Chew\textsuperscript{2} \quad Xucong Zhang\textsuperscript{1}
  \\ 
      \normalsize\textsuperscript{1}Computer Vision Lab, Delft University of Technology \quad 
      \normalsize\textsuperscript{2}Honda Research Institute Japan\quad
}

\newcommand{\methodname}{PolySLGen\xspace}

\begin{document}
\maketitle
\begin{abstract}
Human-like multimodal reaction generation is essential for natural group interactions between humans and embodied AI. However, existing approaches are limited to single-modality or speaking-only responses in dyadic interactions, making them unsuitable for realistic social scenarios. Many also overlook nonverbal cues and complex dynamics of polyadic interactions, both critical for engagement and conversational coherence.
In this work, we present \textbf{\methodname}, an online framework for \textbf{Poly}adic multimodal \textbf{S}peaking and \textbf{L}istening reaction \textbf{Gen}eration. Given past conversation and motion from all participants, \methodname generates a future speaking or listening reaction for a target participant, including speech, body motion, and speaking state score.
To model group interactions effectively, we propose a pose fusion module and a social cue encoder that jointly aggregate motion and social signals from the group. Extensive experiments, along with quantitative and qualitative evaluations, show that \methodname produces contextually appropriate and temporally coherent multimodal reactions, outperforming several adapted and state-of-the-art baselines in motion quality, motion-speech alignment, speaking state prediction, and human-perceived realism. The source code and model are available at \url{https://github.com/zylinzy/PolySLGen}.

\end{abstract}    
\section{Introduction}
\label{sec:introduction}

\begin{figure}[t]
  \centering
  \includegraphics[width=\linewidth]{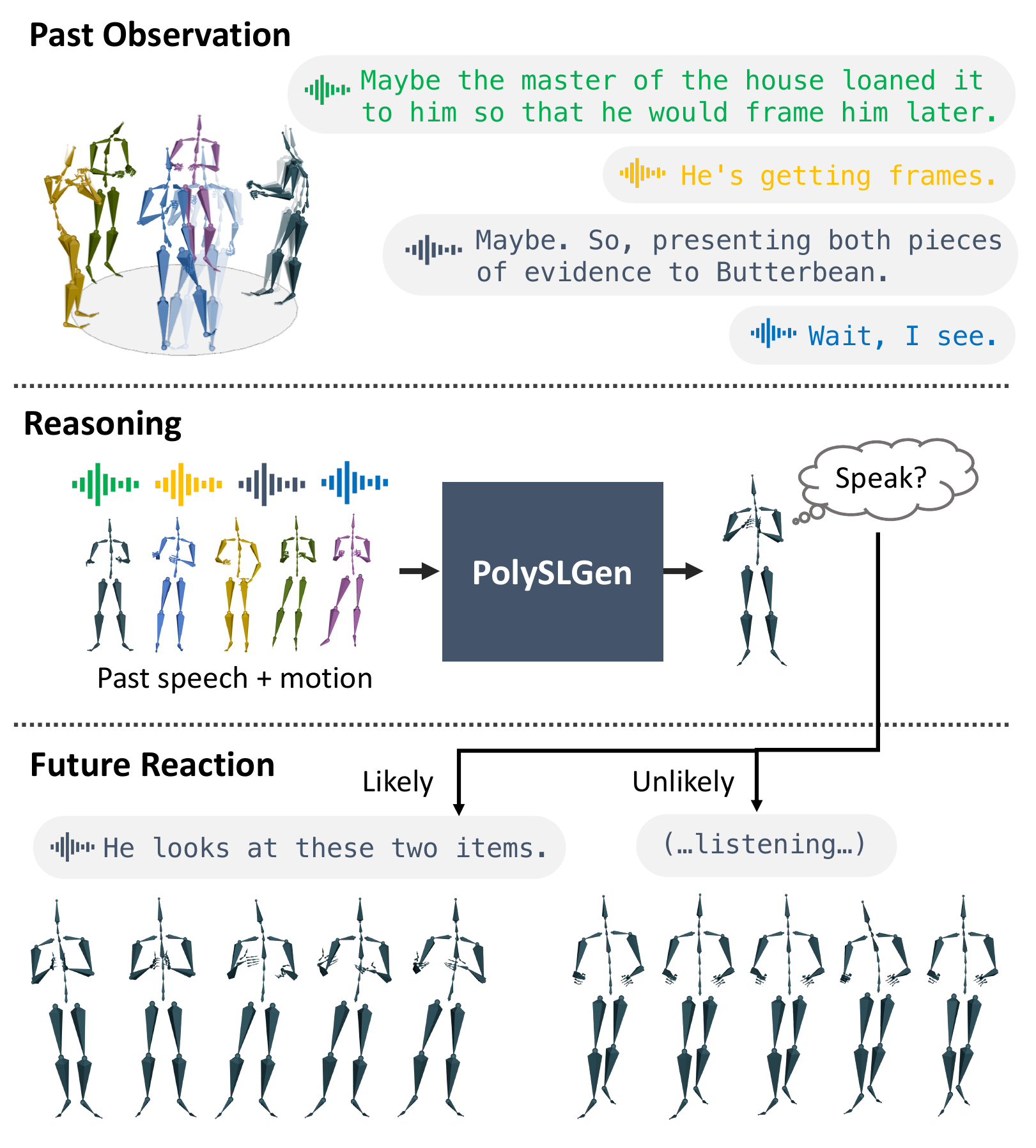}
  \caption{Overview of the online reaction generation task in polyadic interaction. Given past group interactions, including speech and body motion of all participants, \methodname reasons about and generates the future reaction of the target participant (dark blue). The output includes speech, body motion, and a speaking state score (“Speak?”), which serves as an indicator of whether the generated reaction is speaking or listening.
  }
  \label{fig:overview}
\end{figure}


To support natural social interaction, embodied AI systems need to generate responses that coordinate speech and body motion with appropriate conversational turn-taking to signal attention, intention, and manage the flow of conversation \cite{alarcon2021exploring, bisio2014motor, noC4hri2023, russell2025visual}. 
It is then important to model and generate both verbal and non-verbal behaviors to capture complex social dynamics of real social interactions for more efficient and expressive communication \cite{al2012furhat, multimodal_socialdynamics2019, lee2024modeling, li2025towards, multimodal_comm-efficiency2022,noc4empathy2018,noc4deception2019}.

Multimodal Large Language Models (LLMs) have recently shown strong capabilities in motion understanding and generation \cite{zhang2023generating, ribeiro2024motiongpt, zhou2024avatargpt}, social cue interpretation \cite{multimodal&noc_social2024}, and multimodal question answering \cite{zhan2024anygpt, wu2024next, multimodal_roboterror2024, liu2023llava, alayrac2022flamingo}. Leveraging such models enables unified coordination of speech and body motion for more contextually consistent multimodal reactions \cite{jiang2024solami}.

Despite these advances, existing reaction generation approaches face several limitations.
Photorealistic methods produce pixel-perfect outputs but lack the 3D physical grounding required for Embodied AI and Robotics \cite{zhu2025infp, siniukov2025ditailistener}, where spatially consistent skeletal motions are needed for retargeting and control.
Many approaches remain single-modal, generating one modality from another (e.g., motion-to-motion \cite{siyao2024duolando}, text-to-motion \cite{ng2023text2listen}, and audio-to-motion \cite{mughal2024convofusion, ng2024audio}).
Most also rely on future context \cite{siyao2024duolando, ghosh2024remos, xu2024regennet, cen2025ready}, preventing online, causal generation from past observations \cite{li2025towards}. 
Methods without speech generation \cite{tanke2023social, peng2023trajectory, jeong2024multi, xiao2024multi} capture group dynamics only through body motion, limiting meaningful conversational participation.
Recent work, SOLAMI \cite{jiang2024solami}, explores multimodal LLMs for perceiving and generating motion and speech, but it focuses on speaking behavior and remains restricted to dyadic interactions.

Turn-taking, the switch between speaking and listening, is essential for smooth and natural conversation, particularly in polyadic settings \cite{zarkowski2019multi, ishii2022trimodal}. It ensures conversational coherence, maintains engagement, and allows each participant to respond appropriately to the actions of others. However, unified generation of both speaking and listening reactions remains underexplored, as prior works focus on either speaking \cite{yi2022generating, mughal2024convofusion, ng2024audio} or listening responses \cite{jonell2020let, ng2023text2listen, Ng_2022_CVPR}, limiting natural group interactions.

Polyadic interactions are common in real-world scenarios such as group collaboration \cite{ng2023takes}, education \cite{voultsiou2025systematic}, and social support \cite{leff2013computer, kandalaft2013virtual}, but introduce substantial complexity. As the number of participants increases, modeling their interdependent reactions, shaped by speech, gestures, and orientations, becomes more challenging \cite{group_multiagentvsdyadic2023, groupcohesionhri2022, groupengagehri2020, wiemann2017turn, al2012furhat}. Simply extending dyadic architectures to handle polyadic scenarios is computationally inefficient and often fails to capture higher-order dependencies.

To address these challenges, we formulate online multimodal reaction generation as generating the future reaction of a single target participant in polyadic interactions. As shown in \cref{fig:overview}, given past speech and motion from all participants, the model generates future speech, body motion, and a speaking state score that captures turn-taking behavior without enforcing hard transitions. This focus aligns with real-world scenarios, where embodied AI responds to others rather than predicting full group dynamics.

We therefore propose \methodname, a novel framework that generates future reactions for a target participant from past group observations. A pre-trained LLM is adapted for conversation understanding, a pose fusion module aggregates past motions from all participants into compact embeddings, and a social cue encoder captures group-level attention toward the target participant, essential for realistic multi-party interaction. These modules handle variable group sizes and use fixed-length embeddings to preserve LLM context for larger groups.
Combined with speech style generation and speaking state score prediction, \methodname produces coherent multimodal behavior and natural turn-taking.
Extensive experiments and multi-aspect evaluation, including standard objective, social semantic, and human perception metrics, show that \methodname outperforms various baselines, remains robust to missing participants, and enables more realistic polyadic interactions.

In summary, the main contributions of this paper include:
\begin{itemize}
    \item We are the first to propose an online multimodal reaction generation framework with both speaking and listening responses in polyadic interaction settings.
    \item Our approach incorporates speaking state score prediction, allowing the system to dynamically alternate between speaking and listening responses.
    \item Extensive experiments and multi-aspect evaluations show the proposed method outperforms variant baselines in motion, speech, and speaking state prediction.
\end{itemize}

\section{Related Work}

\begin{figure*}[t]
  \centering
  \includegraphics[width=0.92\textwidth]{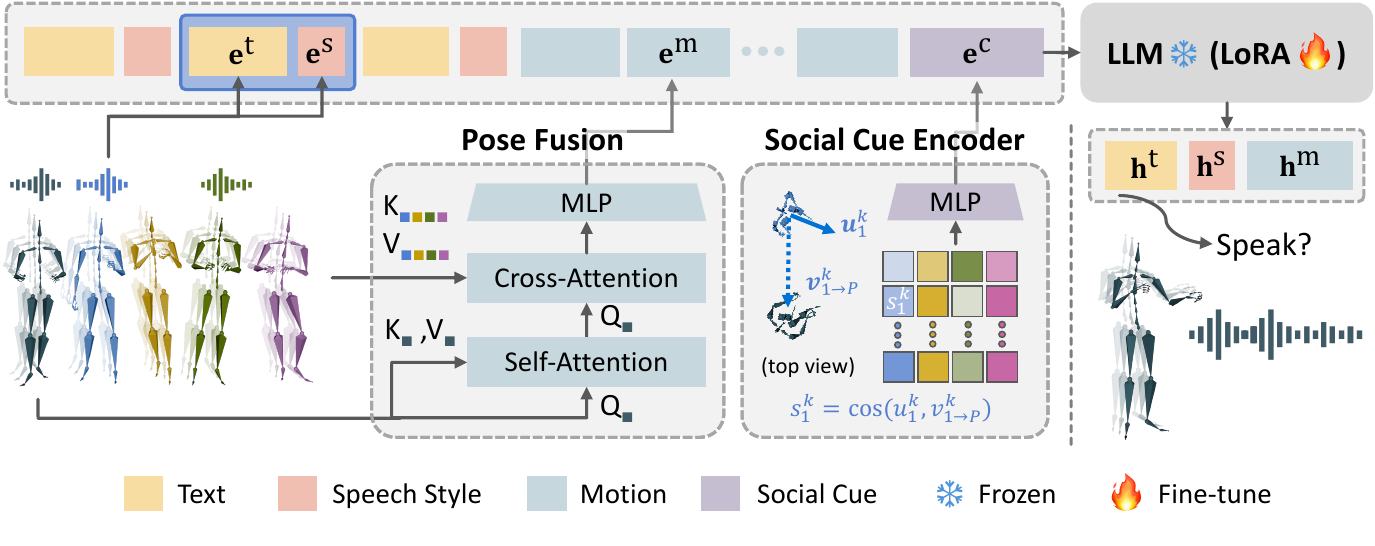}
  \caption{
  Overview of \methodname. Past multimodal group interactions are encoded into text $\mathbf{e}^{\text{t}}$, speech style $\mathbf{e}^{\text{s}}$, motion $\mathbf{e}^{\text{m}}$, and social cue $\mathbf{e}^{\text{c}}$ embeddings as the inputs. The motion embeddings are aggregated by a pose fusion module, and the social cue encoder captures signals from non-target participants’ head orientations. The model generates target participant's future reactions in text, speech style, and motion, through modality-specific decoders from embeddings $\mathbf{h}^{\text{t}}$, $\mathbf{h}^{\text{s}}$, and $\mathbf{h}^{\text{m}}$, along with a speaking state score for turn-taking.
  }
  \label{fig:pipeline}
\end{figure*}

\subsection{Motion and Reaction Generations}
\myheading{Instruction-based Motion Generation}
Recent advancements in text-to-motion generation have been primarily driven by diffusion-based generative models and LLMs.
These approaches have demonstrated effectiveness in generating realistic and semantically coherent human motion conditioned on various modalities, such as natural language descriptions \cite{zhang2024motiondiffuse, Wang_2024_CVPR, hong2025salad, zhang2023generating, ribeiro2024motiongpt, zhang2024large, chen2025language, luo2024m, yu2025socialgen, gong2023tm2d}, audio \cite{zhang2024large, chen2025language, luo2024m, gong2023tm2d}, and speech \cite{zhang2024large, chen2025language, liu2024towards}.

However, most of these approaches require explicit conditioning instructions that specify what motion to generate, which remains largely unaddressed in this line of research. In contrast, our framework aims to generate appropriate future reactions directly based on past observations, without requiring predefined motion instructions.

\myheading{Interactive Motion Generation}
Building upon prior research in conditional motion generation, an emerging direction focuses on reaction generation, which aims to synthesize human motion that dynamically responds to an interacting partner. One line of work explores reaction generation conditioned on verbal communication, such as text or speech, highlighting the strong correlation between spoken language and non-verbal behaviors, such as facial expressions \cite{Ng_2022_CVPR, ng2023text2listen} and full-body gestures \cite{mughal2024convofusion, ng2024audio}. Alternatively, reaction can be generated by conditioning solely on the observed motion of the other participant \cite{siyao2024duolando, ghosh2024remos, xu2024regennet, cen2025ready, park2025unified}. 

Nevertheless, most of the existing methods operate in an offline setting that requires access to both past and future context. This reliance constrains their usefulness to generate seamless reactive motion in real-world scenarios.
Another limitation of current reaction generation methods is their main focus on dyadic interactions, whereas polyadic settings remain underexplored.
While some studies generate future interactive motion in polyadic settings \cite{tanke2023social, peng2023trajectory, jeong2024multi, xiao2024multi}, they predict the joint motion of all participants rather than modeling causal relationships or the reactive behavior of a specific participant.

\myheading{Multimodal Reaction Generation}
To support realistic reaction generation, some recent methods generate both speech and body motion using LLMs. One approach simulates scripted dyadic interactions by retrieving motion from a database using textual input, though this often leads to weak text-motion alignment \cite{cai2024digital}. 
A recent work, SOLAMI \cite{jiang2024solami}, takes both speech and motion conversation of the interacting partner as input to generate a reaction of speech and motion.
However, this method is limited to dyadic interactions, and its performance in polyadic settings remains unclear. More importantly, it focuses only on generating speaking reactions. 
Without modeling of listening reaction, the agent switches to a default motion whenever it is not speaking, which appears unnatural and discontinuous. As a result, the method cannot produce smooth, in-context, embodied behavior for deployment in real-world settings.

Despite recent progress, current reaction generation methods suffer from several key limitations. These include the lack of integrated modeling of both speaking and listening behaviors, insufficient joint modeling of multiple modalities, reliance on offline processing, and a general restriction to dyadic interactions.

\subsection{Multimodal Large Language Models}
With the rise of LLMs \cite{grattafiori2024llama, achiam2023gpt, chatgpt}, efforts have emerged to extend their capabilities beyond text. Vision-Language Models (VLMs) \cite{liu2023llava, alayrac2022flamingo, li2023blip} represent some of the most robust advances, benefiting from training on large-scale image-text datasets. Other approaches achieve cross-modal alignment through learning shared text embeddings \cite{zhu2023languagebind}, tokenizing and fine-tuning across modalities \cite{zhan2024anygpt}, or learning lightweight adapters to avoid modality-specific encoders \cite{han2024onellm, feng2024chatpose}. Still, adapting LLMs to non-text modalities remains challenging due to the scarcity of aligned multimodal data and the inherent differences between data types. LLMs also have difficulty in learning turn-taking behaviors necessary for natural interactions with human users  \cite{umair2024large, arora2025talking}.

In this work, we use an end-to-end adapter learning strategy to learn modality-specific embeddings compatible with the LLM, along with lightweight modality-specific decoders to map LLM outputs back to each modality, without separately pretrained encoders or decoders.
We also model speaking states from the observed group context to better understand and manage turn-taking.

\section{Method}
\label{sec:method}
The architecture is illustrated in \cref{fig:pipeline}. 
The objective of \methodname is to generate the future reaction of a target participant based on past group interaction.
The input consists of past text $\mathcal{X}^{\text{t}}$, speech style $\mathcal{X}^{\text{s}}$, and body and hand motions $\mathcal{X}^{\text{m}}$ from $\text{P}$ participants, where only $\text{P}'$ participants speak in the past conversation. The output includes the target participant's future text $\mathbf{y}^{\text{t}}$, speech style $\mathbf{y}^{\text{s}}$, body and hand motions $\mathbf{y}^{\text{m}}$, and a speaking state score $r$. The overall pipeline is formalized as follows:
\begin{equation}
(\mathbf{y}^{\text{t}}, \mathbf{y}^{\text{s}}, \mathbf{y}^{\text{m}}, r) = \text{\methodname}(\mathcal{X}{^{\text{t}}_{\text{P}'}}, \mathcal{X}{^{\text{s}}_{\text{P}'}}, \mathcal{X}{^{\text{m}}_{\text{P}}}).
\end{equation}

A pre-trained LLM serves as the backbone for interaction reasoning. To integrate speech style and motion, we design dedicated modules to encode and decode embeddings for each modality. To model the group dynamics in the past observation, \methodname introduces a pose fusion module that jointly considers motions from all participants, and a social cue encoder that captures interaction signals from the head orientations of the other participants.

\subsection{Speech}
We take speech conversation in the past $\text{H}$ frames as the observation. 
The raw audio of each participant is first segmented based on utterances using pyannote-audio \cite{bredin2021end}. Then, for each utterance, we convert the audio to text via stable-ts \cite{stable-ts} which uses Whisper \cite{radford2023robust} as the backbone, and speech style feature $\mathbf{x}^{\text{s}} \in \mathbb{R}^{d_{\text{s}}}$ via StyleTTS 2 \cite{li2023styletts}, where $d_{\text{s}}$ denotes the dimensionality of the style feature. This $\mathbf{x}^{\text{s}}$ feature encodes prosodic and emotional characteristics such as speaking rate, intonation, and expressiveness.

To integrate speech style into the LLM input space, we introduce a speech style adapter $\phi^{\text{style}}$
that projects the style feature $\mathbf{x}^{\text{s}}$ into an LLM-compatible embedding as $\mathbf{e}^{\text{s}} = \phi^{\text{style}}(\mathbf{x}^s) \in \mathbb{R}^{d_{\text{llm}}}$,
where $d_{\text{llm}}$ is the dimensionality of the LLM hidden layers.  The converted text can be directly fed into the LLM model to be a text embedding $\mathbf{e}^{\text{t}}$.
Note $\mathcal{X}{^{\text{t}}_{\text{P}'}}=(\mathbf{e}{^t_1}, \mathbf{e}{^t_2}, \dots, \mathbf{e}{^t_{\text{P}'}})^\top \in \mathbb{R}^{\text{P}' \times d_{\text{llm}}}$, and $\mathcal{X}{^{\text{s}}_{\text{P}'}}=(\mathbf{e}{^s_1}, \mathbf{e}{^s_2}, \dots, \mathbf{e}{^s_{\text{P}'}})^\top \in \mathbb{R}^{\text{P}' \times d_{\text{llm}}}$.

Correspondingly, a learnable projection head $f^{\text{style}}$
maps the output speech style embedding $\mathbf{h}^{\text{s}} \in \mathbb{R}^{d_{\text{llm}}}$ back to the original style feature space as $\mathbf{y}^{\text{s}} = f^{\text{style}}(\mathbf{h}^{\text{s}}) \in \mathbb{R}^{d_{\text{s}}}$.
The $\mathbf{y}^{\text{s}}$ and generated text $\mathbf{y}^{\text{t}}$ are used for speech synthesis through the decoder of StyleTTS 2 \cite{li2023styletts}.

\subsection{Motion}
Give the input body and hand motions from the group $ \mathbf{x}^{\text{m}} = (x_1, x_2, \dots, x_{\text{P}})^\top \in \mathbb{R}^{\text{P} \times d_{\text{m}}} $ over the past $\text{H}^{\text{m}}$ frames, one could feed these features sequentially as input. However, the LLM will neglect the coherence between participants due to its causal setting. In addition, the increased motion input from multiple participants consumes context space, which leaves less capacity for linguistic inputs.

To address these challenges, we introduce pose fusion module $ \phi^{\text{motion}}: \mathbb{R}^{\text{P} \times d_{\text{m}}} \rightarrow \mathbb{R}^{d_{\text{llm}}} $ to learn a joint pose embedding from the observed group motion $\mathbf{x}^{\text{m}}$. This joint pose embedding enables the model to capture cross-participant interactions efficiently without increasing the input length as more participants are added.

The pose fusion module $ \phi^{\text{motion}} $ is designed as a hierarchical transformer block, comprising a self-attention layer to encode intra-participant motion dynamics, a cross-attention layer to aggregate inter-participant dependencies, and a multilayer perceptron (MLP) to produce a fused pose embedding $ \mathbf{e}^{\text{m}} \in \mathbb{R}^{d_{\text{llm}}} $. The process is formulated as:
\begin{equation}
\begin{array}{c}
x'^{\text{m}}_{\text{P}} = \text{SelfAtt}(x^{\text{m}}_{\text{P}}),\ \mathbf{x}^{\text{m}}_{\text{others}} = (x_{1}, x_{2}, ..., x_{P-1})^\top, \\[1.5ex]
\mathbf{e}^{\text{m}} = \text{MLP}(\text{CrossAtt}(x'^{\text{m}}_{\text{P}}, \mathbf{x}^{\text{m}}_{\text{others}})), 
\end{array}
\end{equation}
where $ x^{\text{m}}_{\text{P}} $ denotes the pose of the target participant, and $\mathbf{x}^{\text{m}}_{\text{others}}$ denotes the motions of the remaining participants. The resulting embedding $\mathbf{e}^{\text{m}}$ captures temporal dynamics and cross-participant interactions, and provides the LLM with a compact representation of the group’s motions.
For motion generation, a learnable projection head $ f^{\text{motion}}: \mathbb{R}^{d_{\text{llm}}} \rightarrow \mathbb{R}^{d_{\text{m}}} $ maps the output pose embeddings $ \mathbf{h}^{\text{m}} \in \mathbb{R}^{d_{\text{llm}}} $ back to the original motion representation as $ \mathbf{y}^{\text{m}} \in \mathbb{R}^{d_{\text{m}}} $.

\subsection{Social Cue}
In addition to the low-level motion representation, we propose a social cue encoder $\phi^{\text{social}}: \mathbb{R}^{\text{H}^{\text{m}} \times (\text{P}-1) \times d_{\text{rot}}}  \rightarrow \mathbb{R}^{n \times d_{\text{llm}}}$ to learn a high-level social cue embedding from the head orientations, where $d_{\text{rot}}$ is the dimensionality of the head orientation vector, and $n$ is the social cue embedding length.
Inspired by the study of the relationship between turn-taking and visual bodily cues \cite{jokinen2013gaze, kendrick2023turn}, for each non-target participant $i$ at frame $k$, we compute a social cue score $s_i^k$, where $k=-\text{H}^{\text{m}}+1, ..., -1, 0$, and $i = 1, 2, \ldots, \text{P}-1$, to indicate attention toward the target participant.
As shown in \cref{fig:pipeline}, given the head orientation vector $\mathbf{u}_i^k$ and the relative position vector $\mathbf{v}_{i \rightarrow \text{P}}^k$, the score is define as:
\begin{equation}
s_i^k = \cos(\mathbf{u}_i^k, \mathbf{v}_{i \rightarrow \text{P}}^k),
\label{eq:social_cue_score}
\end{equation}
where $s_i^k \in [0,1]$, with higher values representing stronger orientation toward the target. 
The frame-level social signal $\mathbf{s}^k = (s_1^k, s_2^k, \ldots, s_{\text{P}-1}^k)$ is aggregated over all past frames into the temporal social signal $\mathbf{S} = [\mathbf{s}^{-\text{H}^{\text{m}}+1}, \ldots, \mathbf{s}^{-1}, \mathbf{s}^0]^\top \in \mathbb{R}^{\text{H}^{\text{m}} \times (\text{P}-1)}$.
Finally, the social cue embedding $ \mathbf{e}^{\text{c}} \in \mathbb{R}^{n \times d_{\text{llm}}} $ is obtained using a multi-stage MLP that progressively projects and aggregates the temporal social signal $\mathbf{S}$ across time and participants.

\subsection{Speaking State}
\methodname\ also predicts a speaking state score, denoted as $ r \in \mathbb{R} $, reflecting the model's confidence that the target participant should speak. This value is predicted from the first LLM-generated embedding and is formulated as $r = f^{\text{state}}(\mathbf{h}_0)$, where $ f^{\text{state}}: \mathbb{R}^{d_{\text{llm}}} \rightarrow \mathbb{R} $ is an MLP, and $ \mathbf{h}_0 $ denotes the first output embedding. Importantly, the generation of text $\mathbf{y}^{\text{t}}$ and speech style $\mathbf{y}^{\text{s}}$ proceeds independently of $r$. The speaking state score serves as a soft cue to guide turn-taking, rather than enforcing hard transitions, consistent with best practices in human-robot interaction.

\subsection{Loss Function}
The overall loss function is defined as:
\begin{equation}
\mathcal{L}_{\text{total}} = \lambda_{\text{text}} \cdot \mathcal{L}_{\text{text}} + \lambda_{\text{style}} \cdot \mathcal{L}_{\text{style}} + \lambda_{\text{state}} \cdot \mathcal{L}_{\text{state}} + \mathcal{L}_{\text{motion}},
\end{equation}
where $\mathcal{L}_{\text{text}}$, $\mathcal{L}_{\text{style}}$, $\mathcal{L}_{\text{state}}$, and $\mathcal{L}_{\text{motion}}$ represent the losses associated with text token, speech style feature, speaking state score, and motion, respectively. The coefficients $\lambda_{\text{text}}$, $\lambda_{\text{style}}$, and $\lambda_{\text{state}}$ are scalar weights for each loss term.

The textual loss $\mathcal{L}_{\text{text}}$ is computed using a cross-entropy criterion applied to the predicted text tokens. The speech style loss $\mathcal{L}_{\text{style}}$ is defined as the mean squared error between the predicted and ground truth style features. The speaking state score loss $\mathcal{L}_{\text{state}}$ is a binary cross-entropy loss.
The motion loss $\mathcal{L}_{\text{motion}}$ includes a representation-level loss, localized 3D keypoint loss, root position loss, and a regularization term adopted from \cite{rempe2020contact, jang2022motion, kim2024most} to enforce temporal smoothness and ground contact consistency. More details are provided in the supplementary material.
\section{Experiments}

\begin{table*}[t]\caption{Comparison of \methodname with baselines on motion, speech, and speaking state score. SOLAMI is evaluated with LoRA fine-tuning and without pre-training. \textsuperscript{\dag}: Speaking state inferred from generated text. \textsuperscript{*}: Speech synthesized using prompts from test set audio. For the Diversity metric, the value closer to the ground-truth (117.09) is better.
}
\setlength{\tabcolsep}{7pt}
\centering
\begin{tabular}{lccccccccc}
\toprule
\multirow{3}[3]{*}{Method} & \multicolumn{5}{c}{Motion}  & \multicolumn{3}{c}{Speech} & \multicolumn{1}{c}{State} \\ \cmidrule(lr){2-6} \cmidrule(lr){7-9} \cmidrule(lr){10-10}
                           & Root$\downarrow$   & MPJPE$\downarrow$ & \multirow{2}{*}{FID$\downarrow$} & \multirow{2}{*}{Div.$\rightarrow$} & BeatAlign & BERT  & \multirow{2}{*}{WER$\downarrow$} & \multirow{2}{*}{SIM$\uparrow$}  &  \multirow{2}{*}{AP$\uparrow$} \\
                           & (mm)   & (mm) &   &  & Diff.$\downarrow$ & Score$\uparrow$ & &  & \\

\midrule
Random                        & 140.4  & 200.4 & 17.82  & 120.46 & 0.018  & 0.458  & 1.699  & 0.494   & 0.50\textsuperscript{\dag} \\
NN cond.                      & 134.7  & 187.7 & 16.36  & 105.42 & 0.023 & 0.451  & 2.075  & 0.520   & 0.52\textsuperscript{\dag} \\
\midrule
LLM + ConvoFusion \cite{mughal2024convofusion}   & 125.7 & 170.1 & 18.57 & 72.45  &  -     & 0.388 & 13.318  &  - & 0.50\textsuperscript{\dag} \\
LM-L2L Adapted \cite{ng2023text2listen}   & 185.2  & 187.6 & 17.22  & \textbf{116.36} & -  & -  & -  & -  & -\\
SOLAMI \cite{jiang2024solami}             & 188.6  & 180.9 & 14.86 & 100.13 & 0.061 & 0.428 & 1.854 & 0.745\textsuperscript{*} & 0.50\textsuperscript{\dag}\\
Motion Forecast  & 127.0 & 153.8 & 13.93 & 125.53  &  -     & - & -  &  - & - \\
\midrule
Ours  & \textbf{108.7} & \textbf{144.9} & \textbf{12.18} & 113.32 & \textbf{0.007} & \textbf{0.508}  & \textbf{1.436} & \textbf{0.642}  & \textbf{0.67}\\
\bottomrule
\end{tabular}
\label{tab:compare_baseline}
\end{table*}

\subsection{Dataset}
Most existing datasets are unsuitable for our evaluation due to missing modalities \cite{joo2019ssp,raman2022conflab,tanke2023humans,lightcap2021,zhang2024hoi}, lack of 3D pose information \cite{raman2022conflab, northcutt2020egocom}, restrictions to dyadic interactions \cite{lee2019talking, liu2022beat}, or limited multi-person interaction structures \cite{jiang2020coherent, lightcap2021, vendrow2023jrdb}. The most recently released Embody3D \cite{embody3d} offers suitable modalities and polyadic interactions, but after filtering out recordings involving scene interaction, each subject has only about sixteen one-minute-long recordings, which is insufficient to capture stable interaction dynamics.

We therefore adopt the DnD Group Gesture dataset \cite{mughal2024convofusion}, which provides synchronized audio, video, and full 3D body and hand motion for five participants in a tabletop role-playing game. It contains four sessions totaling six hours, with the last reserved for testing. We select the Dungeon Master as the target participant, as their interactions are more frequent and diverse than those of other roles.

\subsection{Implementation Details}
We use Llama3-8B-Instruct \cite{grattafiori2024llama} and apply LoRA fine-tuning \cite{hu2022lora} to the query, key, and value projection layers with rank set to 16, $\alpha$ set to 32, and the dropout rate set to 0.1. The maximum input length for Llama3-8B-Instruct is set to 1,024 tokens.
We use AdamW optimizer \cite{loshchilov2018decoupled} with a batch size of 16, the learning rates 1e-4 for the LLM and 2e-4 for the remaining modules. Training is performed on one NVIDIA A100 GPU.
The model is trained for 20 epochs, with the final model corresponding to the last epoch.
The dimensionality $d_{\text{s}}$, $d_{\text{m}}$, $d_{\text{rot}}$, and $d_{\text{llm}}$, are 256, 327, 6, and 3072. The group size $\text{P}$ is 5 in the DnD dataset. The social cue embedding length $n$ is set to 2. The number of history frames for speech $\text{H}$ and motion $\text{H}^{\text{m}}$ are set to 512 ($\sim$20.48 seconds) and 64 ($\sim$2.56 seconds), respectively. 
More details are in the supplementary material. 

\subsection{Evaluation Metrics}
For motion evaluation, we use the L2 error of the \textbf{Root} joint and Mean Per Joint Position Error (\textbf{MPJPE}) to assess the spatial accuracy of the generated joint positions. Fréchet Inception Distance (\textbf{FID}) and Diversity (\textbf{Div.}) are employed to evaluate the motion similarity to the ground truth motions. 
The difference of the beat alignment (\textbf{BeatAlign Diff.}) evaluates the synchronization between generated speech and the accompanying body and hand movements.

For speech evaluation, we employ \textbf{BERTScore} \cite{bert-score} to measure semantic similarity at the sentence level using contextual embeddings. Word Error Rate (\textbf{WER}) \cite{woodard1982,morris2004} is also reported for lower-level transcription accuracy. To assess voice similarity, we follow prior text-to-speech research \cite{zhang2023speechtokenizer} and compute the cosine similarity (\textbf{SIM}) between speaker embeddings extracted using WavLM-TDNN \cite{chen2022wavlm} from both the generated speech and the ground truth. 

For speaking state score evaluation, we use the Area under the Precision-Recall curve (\textbf{AP}). 
For social semantic evaluation, we use head orientation as a proxy for attention. We measure the target participant’s Mean Angular Error of the head (\textbf{$\text{MAE}_{\text{head}}$}) and the social cue score error for each non-target participant. The social cue score quantifies how much attention the target participant directs toward another one, and the error reflects how much the generated attention differs from the ground truth.

\subsection{Baselines}
\myheading{Random} Return a randomly chosen response segment from the training set as the prediction.

\myheading{NN condition}
Given the input observations, we retrieve the Nearest-Neighbor (NN) in the embedding space from the training set as the output.

\myheading{LLM + ConvoFusion}
We query a language model \cite{grattafiori2024llama} to generate the response based on the past speech conversation in text. The generated text then conditions the co-speech offline full-body motion generation model, ConvoFusion \cite{mughal2024convofusion}, to produce the corresponding motion response.

\myheading{LM-L2L Adapted} 
We extend a state-of-the-art (SOTA) language-model-based dyadic text-to-facial reaction generation method, LM-Learn-to-Listen (LM-L2L) \cite{ng2023text2listen}, to text-to-full-body motion in polyadic interactions.

\myheading{SOLAMI}
We extend the recent reaction generation model SOLAMI \cite{jiang2024solami} from dyadic to polyadic interactions by incorporating additional roles and conditioning on the observed speech and full-body motion of all participants. 
Following its original design, SOLAMI is trained only on speaking reaction data. 
We directly apply LoRA-based instruction fine-tuning for a fair comparison with our method. The LLM embedding layer is fully finetuned to accommodate the newly introduced motion tokens.

\myheading{Motion Forecast}
A variant of \methodname is used as a dedicated polyadic motion forecasting model to benchmark motion quality. It predicts the target participant’s future motion from motion-only past observations.

For all baselines, if no text is generated or the text contains no spoken words, the target participant is considered to be listening. More details are in the supplementary.

\subsection{Comparison to Baselines}
A comparison of \methodname with baselines is presented in \cref{tab:compare_baseline}. 
Across motion metrics, \methodname outperforms all baselines on most error metrics.
We attribute the performance gains to our multimodal framework, which jointly models verbal and non-verbal cues from all participants. Combined with the speaking state score prediction, these components provide a richer interaction context for generating appropriate target reactions.
Notably, several SOTA baselines perform comparably to or worse than Random and NN baselines on motion errors, demonstrating the limitations of adapting methods designed for simpler dyadic settings.
Motion Forecast offers a fairer motion benchmark as adapting SOLAMI \cite{jiang2024solami} to polyadic scenarios may break its dyadic inductive biases. While it improves over SOLAMI \cite{jiang2024solami}, it still underperforms \methodname, highlighting the advantage of jointly modeling motion and conversational context over separate approaches.

For the speech-related metrics, \methodname achieves the best performance across all baselines, even though the SOTA methods LLM+ConvoFusion \cite{mughal2024convofusion} and SOLAMI \cite{jiang2024solami} also use LLMs for text and speech generation.
This suggests that while current LLMs can generate contextually plausible responses, they struggle to generate appropriately timed responses in polyadic interactions. This limitation is reflected in their much lower performance on evaluation metrics such as BERTScore and WER.
In contrast, \methodname jointly considers both the verbal and non-verbal cues from all participants to aid the speech generation.

Regarding the speaking state score prediction, all baselines generate the random chance level prediction, which indicates the extreme difficulty of the task.
In contrast, our \methodname achieves 0.67 AP, better than all baselines with a large margin. This improvement is likely due to the proposed pose fusion and social cue encoding, which provide informative non-verbal cues for the speaking state score prediction.
Note that the higher SIM score of SOLAMI \cite{jiang2024solami} is expected, as it uses the speech style extracted from example audio of the target participant for speech synthesis. However, without explicitly modeling speech style, this can cause misalignment between speech and motion, which explains SOLAMI's lower beat alignment performance. 


For social semantics, as shown in \cref{tab:social_motion}, \methodname achieves the lowest errors on both metrics. Although the improvement in social cue score error is relatively small, it demonstrates that \methodname generates socially coherent behavior and outperforms the two strongest baselines.

\begin{table}[t]
\caption{Evaluation of social semantics. $\text{MAE}_{\text{head}}$ measures head orientation error of the target participant. User 1--4 columns report the social cue score error of each non-target participant.}
\centering
\setlength{\tabcolsep}{2.0pt}
\begin{adjustbox}{width=\linewidth,center}
\begin{tabular}{lccccc}
\toprule
\multirow{2}{*}{Method}  & $\text{MAE}_{\text{head}}\downarrow$ & \multicolumn{4}{c}{Social Cue Score Error$\downarrow$} \\
\cmidrule(lr){3-6}
 & (deg) & User 1 & User 2 & User 3 & User 4 \\
\midrule
LM-L2L Adapted \cite{ng2023text2listen} & 31.7  & 0.35 & 0.14 & 0.31 & 0.20 \\
SOLAMI \cite{jiang2024solami}              & 27.46 & 0.32 & 0.13 & 0.27 & 0.20 \\
\midrule
Ours                                        & \textbf{26.46} & \textbf{0.30} & \textbf{0.12} & \textbf{0.25} & \textbf{0.18} \\
\bottomrule
\end{tabular}
\end{adjustbox}
\label{tab:social_motion}
\end{table}


\subsection{Ablation Studies}

\begin{table*}[t]
\caption{Ablation study on pose fusion and social cue encoder. Diversity closer to the ground truth (117.09) is better.}
\centering
\begin{tabular}{ccccccccccc}
\toprule
\multirow{3}[2]{*}{\makecell[c]{Pose\\Fusion}} & \multirow{3}[2]{*}{\makecell[c]{Social\\Cue}}   & \multicolumn{5}{c}{Motion}  & \multicolumn{3}{c}{Speech} & State \\ \cmidrule(lr){3-7} \cmidrule(lr){8-10} \cmidrule(lr){11-11}
             &        & Root$\downarrow$   & MPJPE$\downarrow$ & \multirow{2}{*}{FID$\downarrow$} & \multirow{2}{*}{Div.$\rightarrow$} & BeatAlign & BERT  & \multirow{2}{*}{WER$\downarrow$} & \multirow{2}{*}{SIM$\uparrow$}  &  \multirow{2}{*}{AP$\uparrow$} \\
                   &        & (mm)   & (mm) &   &  & Diff.$\downarrow$ & Score$\uparrow$ & &  & \\

 \midrule
              \xmark & \xmark & 126.1 & 153.3 & 14.01 & 121.37          & 0.012         & 0.489          & 1.454 & 0.631          & 0.60\\
              \cmark & \xmark & 116.7 & 148.2 & 12.97 & \textbf{118.34} & \textbf{0.007}  & \textbf{0.511} & 1.447 & \textbf{0.646} & 0.66 \\
              \xmark & \cmark & 124.6 & 152.5 & 13.53 & 123.19          & 0.009         & 0.474          & 1.715 & 0.625          & 0.59 \\
\midrule              
              \cmark & \cmark & \textbf{108.7} & \textbf{144.9} & \textbf{12.18} & 113.32 & \textbf{0.007} & 0.508  & \textbf{1.436} & 0.642  & \textbf{0.67}\\
\bottomrule
\end{tabular}
\label{tab:ablation_components}
\end{table*}

\myheading{Model Components}
We first assess the incremental contribution of the proposed pose fusion and social cue encoder in \cref{tab:ablation_components}.
The first row shows the model without both modules. By adding the pose fusion into the model, we observe significant improvements over motion metrics, semantic and acoustic similarity in speech, and speaking state score AP. 
These improvements could be from the joint modeling of motions from multiple participants before sending them to the LLM for reasoning. 

Adding the social cue encoder alone does not yield consistent improvements (third row), possibly because it operates on relatively high-level features without sufficient grounding in motion dynamics. However, when combined with pose fusion (last row), it provides additional gains in motion quality and speaking state score AP, with a minor trade-off in WER and BERTScore.
These results suggest that multi-person motion understanding provides a solid foundation for social behavior generation in \methodname, and that high-level social cues are most effective when integrated with low-level motion representations.

\begin{table}[t]
\caption{Effect of motion observation. Rows 1–3 progressively add motions from speaking and non-speaking participants. Diversity closer to ground-truth (117.09) is better.}
\centering
\setlength{\tabcolsep}{2.5pt}
\begin{adjustbox}{width=\linewidth,center}
\begin{tabular}{lcccc}
\toprule
\multirow{1}{*}{Method} 
                           & Root$\downarrow$   & MPJPE$\downarrow$ & \multirow{1}{*}{FID$\downarrow$} & \multirow{1}{*}{Div.$\rightarrow$}\\ 

\midrule
w/o motion observation & 165.1  & 202.5 & 19.42  & 124.89 \\
+ verbal motion       &  157.6  & 175.5  &  16.36  & 123.18 \\
+ non-verbal motion & \textbf{126.1} & \textbf{153.3 }& 14.01 & \textbf{121.37}  \\
\bottomrule
\end{tabular}
\end{adjustbox}
\label{tab:motion_obs}
\end{table}

\myheading{Motion Observation}
One of the key contributions of \methodname is the inclusion of motions from both speaking and non-speaking participants in the past observation, unlike most existing works.
To investigate the effect of motion observation, we conduct incremental experiments shown in \cref{tab:motion_obs}.
Starting from a simple baseline with past observation only includes conversation in speech (first row), we find that adding motions from speaking participants already improves the performance, and further including motions from non-speaking participants yields the best results across all motion metrics. The significant performance improvement demonstrates that motions from both speaking and non-speaking participants provide valuable contextual and interpersonal information.

\myheading{Robustness to Missing Participants}
In real-world polyadic interactions, participant modalities may be missing due to occlusion, sensor failure, or incomplete detection, which pose challenges for generating coherent group behaviors. \methodname handles such cases naturally since its architecture supports a variable number of participants, and missing motion or social cues can be zero-padded or ignored during inference. We evaluate robustness by randomly omitting up to 1--3 non-target participants from the DnD dataset. 
As shown in \Cref{tab:robustness}, removing a single participant ($\leq$1) increases MPJPE and FID, indicating reliance on full group context. From $\leq$1 to $\leq$2 missing participants, speaking state AP drops further, while additional removals ($\leq$3) have minor impact. This is likely because only two non-target participants are typically active. Importantly, even under such incomplete input conditions, \methodname still slightly outperforms SOLAMI \cite{jiang2024solami}, which has full group observation. This shows \methodname is robust and reliable for real-world deployment.

\begin{table}[t]
\caption{Impact of missing participants during inference. To simulate real-world scenarios, we randomly remove up to three non-target participants from the DnD Group Gesture dataset.}
\centering
\setlength{\tabcolsep}{4.8pt}
\begin{adjustbox}{width=\linewidth,center}
\begin{tabular}{lcccc}
\toprule
Participants & MPJPE$\downarrow$ & FID$\downarrow$ & BERTScore$\uparrow$  & State AP$\uparrow$ \\
\midrule
0 Missing   & \textbf{144.9} & \textbf{12.18} & 0.508 & \textbf{0.67} \\
$\leq$1 Missing & 176.3 & 15.43 & \textbf{0.509}  & 0.66 \\
$\leq$2 Missing & 183.2 & 16.09 & 0.506  & 0.65 \\ 
$\leq$3 Missing & 189.6 & 16.73 & 0.502  & 0.65 \\

\bottomrule
\end{tabular}
\end{adjustbox}
\label{tab:robustness}
\end{table}

\subsection{Qualitative Results}

\begin{figure*}[t]
  \centering
  \includegraphics[width=0.87\textwidth]{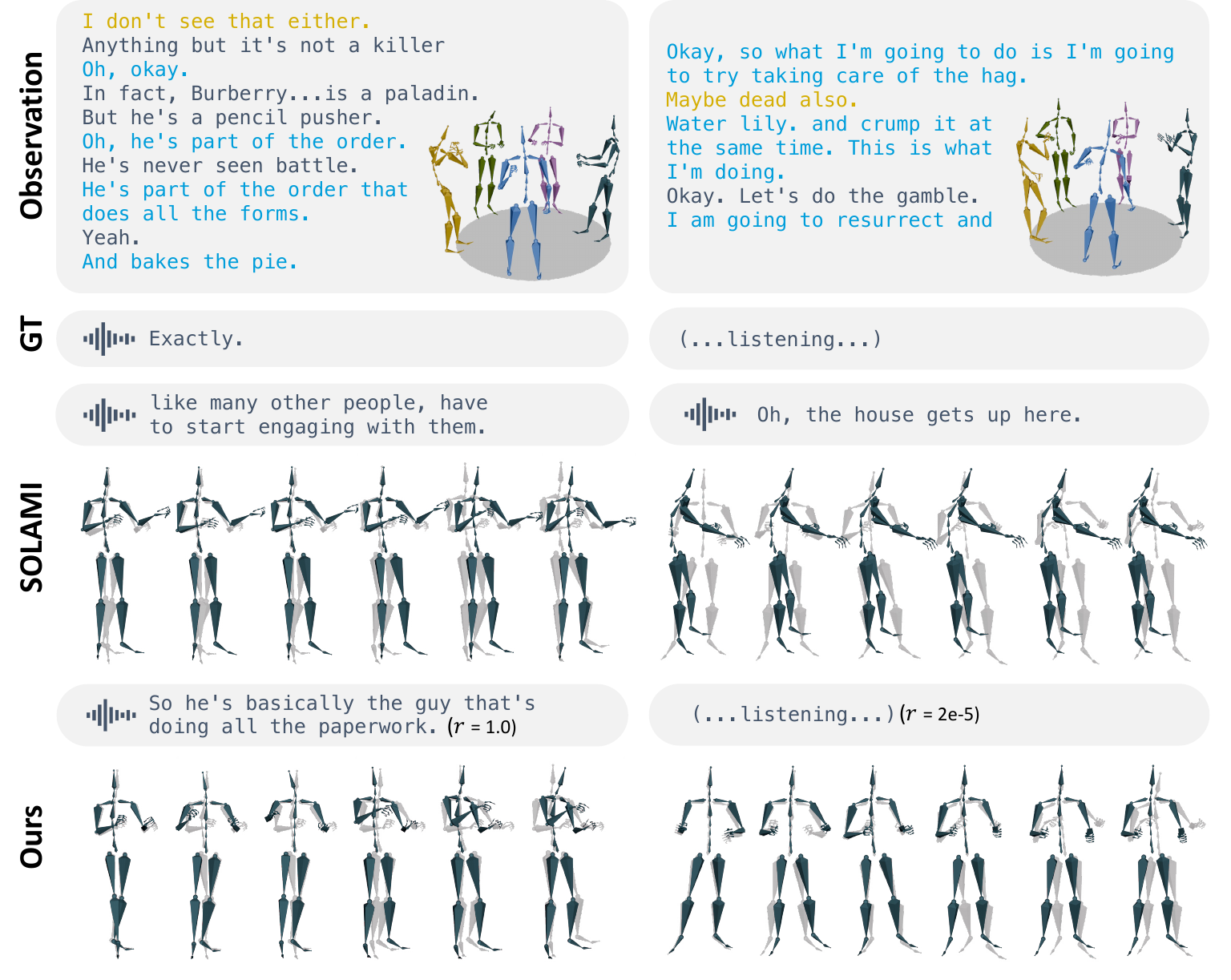}
  \caption{Visual comparison between \methodname and SOLAMI. The first column presents speaking reactions, and the second column shows listening reactions. The top row contains input observations, followed by the ground truth and outputs from SOLAMI and \methodname. The predicted speaking state score is denoted by $r$, and different colors correspond to different speakers. Note that we overlay the ground truth motion (in gray) with the generated outputs.}
  \label{fig:vis_compare}
\end{figure*}

\myheading{Visual Comparison}
We present visual comparisons between \methodname and the competitive baseline SOLAMI \cite{jiang2024solami} in \cref{fig:vis_compare}. Compared to SOLAMI, \methodname effectively captures transitions between speaking and listening states, and the generated motions align more closely with the ground truth. Although the speech semantics differ from the ground truth, this is an expected outcome given the improvisational nature of Dungeons \& Dragons setting. Overall, \methodname generates speech that remains contextually coherent with the conversation, while the output of SOLAMI often appears out of context.

\myheading{User Study}
We conducted a user study as qualitative support for the quantitative results. Five sessions were randomly selected, and reactions were generated using SOLAMI \cite{jiang2024solami} and our \methodname. Twenty-three participants rated each video on motion coherence, motion continuity, speech semantics, speech tone, and overall naturalness using a 5-point Likert scale (1 = poor, 5 = excellent). As shown in \cref{tab:user_study}, \methodname is consistently preferred over SOLAMI, with the largest improvement in motion continuity. By modeling the past motion of the target participant, \methodname produces smooth transitions from past to future. These results also confirm that SOTA dyadic methods do not extend effectively to polyadic interaction scenarios. More details are in the supplementary.

\begin{table}[t]
\caption{User study results (mean ± std) comparing SOLAMI and \methodname. Higher scores indicate better perceived quality.}
\setlength{\tabcolsep}{2pt}
\renewcommand{\arraystretch}{0.8}
\centering
\begin{adjustbox}{width=0.8\linewidth,center}
\begin{tabular}{ccc}
\toprule
Aspect  & SOLAMI \cite{jiang2024solami} & \methodname \\
\midrule
Motion Coherence  & 2.7 $\pm$ 1.2 & \textbf{3.6 $\pm$ 1.1} \\
Motion Continuity  & 2.4 $\pm$ 1.2 & \textbf{3.8 $\pm$ 1.2} \\
\midrule
Speech Semantics  & 2.3 $\pm$ 1.1 & \textbf{3.6 $\pm$ 1.3} \\
Speech Tone & 3.1 $\pm$ 1.2 & \textbf{3.9 $\pm$ 1.0} \\
\midrule
Overall & 2.5 $\pm$ 1.0 & \textbf{3.4 $\pm$ 1.1} \\
\bottomrule
\end{tabular}
\end{adjustbox}
\label{tab:user_study}
\end{table}
\section{Conclusion}
\label{sec:conclusion}
We propose \methodname, a unified reaction generation framework for speaking and listening reactions with explicit speaking state score prediction. Designed for polyadic interactions, it jointly processes all participants via the proposed pose fusion module and social cue encoder to capture group dynamics.
Extensive experiments confirm the task’s challenges and show that \methodname outperforms adapted baselines across most metrics.
\\
\textbf{Limitations and Future Works.} Speaking state prediction remains challenging due to task complexity. Further improvements may come from better transcript understanding and richer social signals.
The limited domain of the DnD Group Gesture dataset motivates more diverse datasets with natural interactions for comprehensive evaluation.
Last, while the online setting supports real-time inference, \methodname is not yet fully real-time, running at $\sim$5 FPS on an A100 GPU, with $\sim$82\% of the runtime spent on the LLM. Future deployment could optimize LLM inference to improve throughput and reduce latency.
{
    \small
    \bibliographystyle{ieeenat_fullname}
    \bibliography{main}

\begin{thebibliography}{97}
\providecommand{\natexlab}[1]{#1}
\providecommand{\url}[1]{\texttt{#1}}
\expandafter\ifx\csname urlstyle\endcsname\relax
  \providecommand{\doi}[1]{doi: #1}\else
  \providecommand{\doi}{doi: \begingroup \urlstyle{rm}\Url}\fi

\bibitem[Achiam et~al.(2023)Achiam, Adler, Agarwal, Ahmad, Akkaya, Aleman, Almeida, Altenschmidt, Altman, Anadkat, et~al.]{achiam2023gpt}
Josh Achiam, Steven Adler, Sandhini Agarwal, Lama Ahmad, Ilge Akkaya, Florencia~Leoni Aleman, Diogo Almeida, Janko Altenschmidt, Sam Altman, Shyamal Anadkat, et~al.
\newblock Gpt-4 technical report.
\newblock \emph{arXiv preprint arXiv:2303.08774}, 2023.

\bibitem[Al~Moubayed et~al.(2012)Al~Moubayed, Beskow, Skantze, and Granstr{\"o}m]{al2012furhat}
Samer Al~Moubayed, Jonas Beskow, Gabriel Skantze, and Bj{\"o}rn Granstr{\"o}m.
\newblock Furhat: a back-projected human-like robot head for multiparty human-machine interaction.
\newblock In \emph{Cognitive behavioural systems: COST 2102 international training school, dresden, Germany, february 21-26, 2011, revised selected papers}, pages 114--130. Springer, 2012.

\bibitem[Alarcon et~al.(2021)Alarcon, Gibson, Jessup, and Capiola]{alarcon2021exploring}
Gene~M Alarcon, Anthony~M Gibson, Sarah~A Jessup, and August Capiola.
\newblock Exploring the differential effects of trust violations in human-human and human-robot interactions.
\newblock \emph{Applied Ergonomics}, 93:\penalty0 103350, 2021.

\bibitem[Alayrac et~al.(2022)Alayrac, Donahue, Luc, Miech, Barr, Hasson, Lenc, Mensch, Millican, Reynolds, et~al.]{alayrac2022flamingo}
Jean-Baptiste Alayrac, Jeff Donahue, Pauline Luc, Antoine Miech, Iain Barr, Yana Hasson, Karel Lenc, Arthur Mensch, Katherine Millican, Malcolm Reynolds, et~al.
\newblock Flamingo: a visual language model for few-shot learning.
\newblock \emph{Advances in neural information processing systems}, 35:\penalty0 23716--23736, 2022.

\bibitem[Arora et~al.(2025)Arora, Lu, Chiu, Pang, and Watanabe]{arora2025talking}
Siddhant Arora, Zhiyun Lu, Chung-Cheng Chiu, Ruoming Pang, and Shinji Watanabe.
\newblock Talking turns: Benchmarking audio foundation models on turn-taking dynamics.
\newblock \emph{ICLR}, 2025.

\bibitem[Bisio et~al.(2014)Bisio, Sciutti, Nori, Metta, Fadiga, Sandini, and Pozzo]{bisio2014motor}
Ambra Bisio, Alessandra Sciutti, Francesco Nori, Giorgio Metta, Luciano Fadiga, Giulio Sandini, and Thierry Pozzo.
\newblock Motor contagion during human-human and human-robot interaction.
\newblock \emph{PloS one}, 9\penalty0 (8):\penalty0 e106172, 2014.

\bibitem[Bredin and Laurent(2021)]{bredin2021end}
Herv{\'e} Bredin and Antoine Laurent.
\newblock End-to-end speaker segmentation for overlap-aware resegmentation.
\newblock \emph{arXiv preprint arXiv:2104.04045}, 2021.

\bibitem[Cai et~al.(2024)Cai, Jiang, Qing, Guo, Zhang, Lin, Mei, Wei, Wang, Yin, et~al.]{cai2024digital}
Zhongang Cai, Jianping Jiang, Zhongfei Qing, Xinying Guo, Mingyuan Zhang, Zhengyu Lin, Haiyi Mei, Chen Wei, Ruisi Wang, Wanqi Yin, et~al.
\newblock Digital life project: Autonomous 3d characters with social intelligence.
\newblock In \emph{Proceedings of the IEEE/CVF Conference on Computer Vision and Pattern Recognition}, pages 582--592, 2024.

\bibitem[Cen et~al.(2025)Cen, Pi, Peng, Shuai, Shen, Bao, Zhou, and Hu]{cen2025ready}
Zhi Cen, Huaijin Pi, Sida Peng, Qing Shuai, Yujun Shen, Hujun Bao, Xiaowei Zhou, and Ruizhen Hu.
\newblock Ready-to-react: Online reaction policy for two-character interaction generation.
\newblock \emph{arXiv preprint arXiv:2502.20370}, 2025.

\bibitem[Chen et~al.(2025)Chen, Zhang, Lakshmikanth, Fang, Shao, Wetzstein, Fei-Fei, and Adeli]{chen2025language}
Changan Chen, Juze Zhang, Shrinidhi~K Lakshmikanth, Yusu Fang, Ruizhi Shao, Gordon Wetzstein, Li Fei-Fei, and Ehsan Adeli.
\newblock The language of motion: Unifying verbal and non-verbal language of 3d human motion.
\newblock In \emph{Proceedings of the Computer Vision and Pattern Recognition Conference}, pages 6200--6211, 2025.

\bibitem[Chen et~al.(2022)Chen, Wang, Chen, Wu, Liu, Chen, Li, Kanda, Yoshioka, Xiao, et~al.]{chen2022wavlm}
Sanyuan Chen, Chengyi Wang, Zhengyang Chen, Yu Wu, Shujie Liu, Zhuo Chen, Jinyu Li, Naoyuki Kanda, Takuya Yoshioka, Xiong Xiao, et~al.
\newblock Wavlm: Large-scale self-supervised pre-training for full stack speech processing.
\newblock \emph{IEEE Journal of Selected Topics in Signal Processing}, 16\penalty0 (6):\penalty0 1505--1518, 2022.

\bibitem[Dahiya et~al.(2023)Dahiya, Aroyo, Dautenhahn, and Smith]{group_multiagentvsdyadic2023}
Abhinav Dahiya, Alexander~M. Aroyo, Kerstin Dautenhahn, and Stephen~L. Smith.
\newblock A survey of multi-agent human–robot interaction systems.
\newblock \emph{Robotics and Autonomous Systems}, 161:\penalty0 104335, 2023.

\bibitem[Feng et~al.(2024)Feng, Lin, Dwivedi, Sun, Patel, and Black]{feng2024chatpose}
Yao Feng, Jing Lin, Sai~Kumar Dwivedi, Yu Sun, Priyanka Patel, and Michael~J Black.
\newblock Chatpose: Chatting about 3d human pose.
\newblock In \emph{Proceedings of the IEEE/CVF conference on computer vision and pattern recognition}, pages 2093--2103, 2024.

\bibitem[Ghosh et~al.(2024)Ghosh, Dabral, Golyanik, Theobalt, and Slusallek]{ghosh2024remos}
Anindita Ghosh, Rishabh Dabral, Vladislav Golyanik, Christian Theobalt, and Philipp Slusallek.
\newblock Remos: 3d motion-conditioned reaction synthesis for two-person interactions.
\newblock In \emph{European Conference on Computer Vision (ECCV)}, 2024.

\bibitem[Gong et~al.(2023)Gong, Lian, Chang, Guo, Jiang, Zuo, Mi, and Wang]{gong2023tm2d}
Kehong Gong, Dongze Lian, Heng Chang, Chuan Guo, Zihang Jiang, Xinxin Zuo, Michael~Bi Mi, and Xinchao Wang.
\newblock Tm2d: Bimodality driven 3d dance generation via music-text integration.
\newblock In \emph{Proceedings of the IEEE/CVF International Conference on Computer Vision}, pages 9942--9952, 2023.

\bibitem[Grattafiori et~al.(2024)Grattafiori, Dubey, Jauhri, Pandey, Kadian, Al-Dahle, Letman, Mathur, Schelten, Vaughan, et~al.]{grattafiori2024llama}
Aaron Grattafiori, Abhimanyu Dubey, Abhinav Jauhri, Abhinav Pandey, Abhishek Kadian, Ahmad Al-Dahle, Aiesha Letman, Akhil Mathur, Alan Schelten, Alex Vaughan, et~al.
\newblock The llama 3 herd of models.
\newblock \emph{arXiv preprint arXiv:2407.21783}, 2024.

\bibitem[Guo et~al.(2022{\natexlab{a}})Guo, Zou, Zuo, Wang, Ji, Li, and Cheng]{guo2022generating}
Chuan Guo, Shihao Zou, Xinxin Zuo, Sen Wang, Wei Ji, Xingyu Li, and Li Cheng.
\newblock Generating diverse and natural 3d human motions from text.
\newblock In \emph{Proceedings of the IEEE/CVF conference on computer vision and pattern recognition}, pages 5152--5161, 2022{\natexlab{a}}.

\bibitem[Guo et~al.(2022{\natexlab{b}})Guo, Zuo, Wang, and Cheng]{guo2022tm2t}
Chuan Guo, Xinxin Zuo, Sen Wang, and Li Cheng.
\newblock Tm2t: Stochastic and tokenized modeling for the reciprocal generation of 3d human motions and texts.
\newblock In \emph{European Conference on Computer Vision}, pages 580--597. Springer, 2022{\natexlab{b}}.

\bibitem[Han et~al.(2024)Han, Gong, Zhang, Wang, Zhang, Lin, Qiao, Gao, and Yue]{han2024onellm}
Jiaming Han, Kaixiong Gong, Yiyuan Zhang, Jiaqi Wang, Kaipeng Zhang, Dahua Lin, Yu Qiao, Peng Gao, and Xiangyu Yue.
\newblock Onellm: One framework to align all modalities with language.
\newblock In \emph{Proceedings of the IEEE/CVF Conference on Computer Vision and Pattern Recognition}, pages 26584--26595, 2024.

\bibitem[Hong et~al.(2025)Hong, Kim, Yoon, Nam, Cha, and Noh]{hong2025salad}
Seokhyeon Hong, Chaelin Kim, Serin Yoon, Junghyun Nam, Sihun Cha, and Junyong Noh.
\newblock Salad: Skeleton-aware latent diffusion for text-driven motion generation and editing.
\newblock In \emph{Proceedings of the Computer Vision and Pattern Recognition Conference}, pages 7158--7168, 2025.

\bibitem[Hu et~al.(2022)Hu, Shen, Wallis, Allen-Zhu, Li, Wang, Wang, Chen, et~al.]{hu2022lora}
Edward~J Hu, Yelong Shen, Phillip Wallis, Zeyuan Allen-Zhu, Yuanzhi Li, Shean Wang, Lu Wang, Weizhu Chen, et~al.
\newblock Lora: Low-rank adaptation of large language models.
\newblock \emph{ICLR}, 1\penalty0 (2):\penalty0 3, 2022.

\bibitem[Ishii et~al.(2022)Ishii, Ren, Muszynski, and Morency]{ishii2022trimodal}
Ryo Ishii, Xutong Ren, Michal Muszynski, and Louis-Philippe Morency.
\newblock Trimodal prediction of speaking and listening willingness to help improve turn-changing modeling.
\newblock \emph{Frontiers in Psychology}, 13:\penalty0 774547, 2022.

\bibitem[Jang et~al.(2022)Jang, Park, and Lee]{jang2022motion}
Deok-Kyeong Jang, Soomin Park, and Sung-Hee Lee.
\newblock Motion puzzle: Arbitrary motion style transfer by body part.
\newblock \emph{ACM Transactions on Graphics (TOG)}, 41\penalty0 (3):\penalty0 1--16, 2022.

\bibitem[Jeong et~al.(2024)Jeong, Park, and Yoon]{jeong2024multi}
Jaewoo Jeong, Daehee Park, and Kuk-Jin Yoon.
\newblock Multi-agent long-term 3d human pose forecasting via interaction-aware trajectory conditioning.
\newblock In \emph{Proceedings of the IEEE/CVF Conference on Computer Vision and Pattern Recognition}, pages 1617--1628, 2024.

\bibitem[Jiang et~al.(2024)Jiang, Xiao, Lin, Zhang, Ren, Gao, Lin, Cai, Yang, and Liu]{jiang2024solami}
Jianping Jiang, Weiye Xiao, Zhengyu Lin, Huaizhong Zhang, Tianxiang Ren, Yang Gao, Zhiqian Lin, Zhongang Cai, Lei Yang, and Ziwei Liu.
\newblock Solami: Social vision-language-action modeling for immersive interaction with 3d autonomous characters.
\newblock \emph{arXiv preprint arXiv:2412.00174}, 2024.

\bibitem[Jiang et~al.(2020)Jiang, Kolotouros, Pavlakos, Zhou, and Daniilidis]{jiang2020coherent}
Wen Jiang, Nikos Kolotouros, Georgios Pavlakos, Xiaowei Zhou, and Kostas Daniilidis.
\newblock Coherent reconstruction of multiple humans from a single image.
\newblock In \emph{Proceedings of the IEEE/CVF conference on computer vision and pattern recognition}, pages 5579--5588, 2020.

\bibitem[Jokinen et~al.(2013)Jokinen, Furukawa, Nishida, and Yamamoto]{jokinen2013gaze}
Kristiina Jokinen, Hirohisa Furukawa, Masafumi Nishida, and Seiichi Yamamoto.
\newblock Gaze and turn-taking behavior in casual conversational interactions.
\newblock \emph{ACM Transactions on Interactive Intelligent Systems (TiiS)}, 3\penalty0 (2):\penalty0 1--30, 2013.

\bibitem[Jonell et~al.(2020)Jonell, Kucherenko, Henter, and Beskow]{jonell2020let}
Patrik Jonell, Taras Kucherenko, Gustav~Eje Henter, and Jonas Beskow.
\newblock Let's face it: Probabilistic multi-modal interlocutor-aware generation of facial gestures in dyadic settings.
\newblock In \emph{Proceedings of the 20th ACM International Conference on Intelligent Virtual Agents}, pages 1--8, 2020.

\bibitem[Joo et~al.(2019)Joo, Simon, Cikara, and Sheikh]{joo2019ssp}
Hanbyul Joo, Tomas Simon, Mina Cikara, and Yaser Sheikh.
\newblock Towards social artificial intelligence: Nonverbal social signal prediction in a triadic interaction.
\newblock In \emph{CVPR}, 2019.

\bibitem[Kandalaft et~al.(2013)Kandalaft, Didehbani, Krawczyk, Allen, and Chapman]{kandalaft2013virtual}
Michelle~R Kandalaft, Nyaz Didehbani, Daniel~C Krawczyk, Tandra~T Allen, and Sandra~B Chapman.
\newblock Virtual reality social cognition training for young adults with high-functioning autism.
\newblock \emph{Journal of autism and developmental disorders}, 43:\penalty0 34--44, 2013.

\bibitem[Kendrick et~al.(2023)Kendrick, Holler, and Levinson]{kendrick2023turn}
Kobin~H Kendrick, Judith Holler, and Stephen~C Levinson.
\newblock Turn-taking in human face-to-face interaction is multimodal: gaze direction and manual gestures aid the coordination of turn transitions.
\newblock \emph{Philosophical transactions of the royal society B}, 378\penalty0 (1875):\penalty0 20210473, 2023.

\bibitem[Kim et~al.(2024)Kim, Kim, Chang, and Choi]{kim2024most}
Boeun Kim, Jungho Kim, Hyung~Jin Chang, and Jin~Young Choi.
\newblock Most: Motion style transformer between diverse action contents.
\newblock In \emph{Proceedings of the IEEE/CVF Conference on Computer Vision and Pattern Recognition}, pages 1705--1714, 2024.

\bibitem[Lee et~al.(2019)Lee, Deng, Ma, Shiratori, Srinivasa, and Sheikh]{lee2019talking}
Gilwoo Lee, Zhiwei Deng, Shugao Ma, Takaaki Shiratori, Siddhartha~S Srinivasa, and Yaser Sheikh.
\newblock Talking with hands 16.2 m: A large-scale dataset of synchronized body-finger motion and audio for conversational motion analysis and synthesis.
\newblock In \emph{Proceedings of the IEEE/CVF International Conference on Computer Vision}, pages 763--772, 2019.

\bibitem[Lee et~al.(2024{\natexlab{a}})Lee, Lai, Ryan, Boote, and Rehg]{lee2024modeling}
Sangmin Lee, Bolin Lai, Fiona Ryan, Bikram Boote, and James~M Rehg.
\newblock Modeling multimodal social interactions: new challenges and baselines with densely aligned representations.
\newblock In \emph{Proceedings of the IEEE/CVF Conference on Computer Vision and Pattern Recognition}, pages 14585--14595, 2024{\natexlab{a}}.

\bibitem[Lee et~al.(2024{\natexlab{b}})Lee, Lai, Ryan, Boote, and Rehg]{multimodal&noc_social2024}
Sangmin Lee, Bolin Lai, Fiona Ryan, Bikram Boote, and James~M. Rehg.
\newblock Modeling multimodal social interactions: New challenges and baselines with densely aligned representations.
\newblock In \emph{2024 IEEE/CVF Conference on Computer Vision and Pattern Recognition (CVPR)}, pages 14585--14595, 2024{\natexlab{b}}.

\bibitem[Leff et~al.(2013)Leff, Williams, Huckvale, Arbuthnot, and Leff]{leff2013computer}
Julian Leff, Geoffrey Williams, Mark~A Huckvale, Maurice Arbuthnot, and Alex~P Leff.
\newblock Computer-assisted therapy for medication-resistant auditory hallucinations: proof-of-concept study.
\newblock \emph{The British Journal of Psychiatry}, 202\penalty0 (6):\penalty0 428--433, 2013.

\bibitem[Li et~al.(2023{\natexlab{a}})Li, Li, Savarese, and Hoi]{li2023blip}
Junnan Li, Dongxu Li, Silvio Savarese, and Steven Hoi.
\newblock Blip-2: Bootstrapping language-image pre-training with frozen image encoders and large language models.
\newblock In \emph{International conference on machine learning}, pages 19730--19742. PMLR, 2023{\natexlab{a}}.

\bibitem[Li et~al.(2025)Li, Deng, Lai, Pian, Rehg, and Tian]{li2025towards}
Xinpeng Li, Shijian Deng, Bolin Lai, Weiguo Pian, James~M Rehg, and Yapeng Tian.
\newblock Towards online multi-modal social interaction understanding.
\newblock \emph{arXiv preprint arXiv:2503.19851}, 2025.

\bibitem[Li et~al.(2023{\natexlab{b}})Li, Han, Raghavan, Mischler, and Mesgarani]{li2023styletts}
Yinghao~Aaron Li, Cong Han, Vinay Raghavan, Gavin Mischler, and Nima Mesgarani.
\newblock Styletts 2: Towards human-level text-to-speech through style diffusion and adversarial training with large speech language models.
\newblock \emph{NeurIPS}, 36:\penalty0 19594--19621, 2023{\natexlab{b}}.

\bibitem[Liu et~al.(2022)Liu, Zhu, Iwamoto, Peng, Li, Zhou, Bozkurt, and Zheng]{liu2022beat}
Haiyang Liu, Zihao Zhu, Naoya Iwamoto, Yichen Peng, Zhengqing Li, You Zhou, Elif Bozkurt, and Bo Zheng.
\newblock Beat: A large-scale semantic and emotional multi-modal dataset for conversational gestures synthesis.
\newblock In \emph{European conference on computer vision}, pages 612--630. Springer, 2022.

\bibitem[Liu et~al.(2023)Liu, Li, Wu, and Lee]{liu2023llava}
Haotian Liu, Chunyuan Li, Qingyang Wu, and Yong~Jae Lee.
\newblock Visual instruction tuning.
\newblock In \emph{NeurIPS}, 2023.

\bibitem[Liu et~al.(2024)Liu, Cao, Wen, Jiang, and Ding]{liu2024towards}
Yifei Liu, Qiong Cao, Yandong Wen, Huaiguang Jiang, and Changxing Ding.
\newblock Towards variable and coordinated holistic co-speech motion generation.
\newblock In \emph{Proceedings of the IEEE/CVF Conference on Computer Vision and Pattern Recognition}, pages 1566--1576, 2024.

\bibitem[Loshchilov and Hutter(2018)]{loshchilov2018decoupled}
Ilya Loshchilov and Frank Hutter.
\newblock Decoupled weight decay regularization.
\newblock In \emph{International Conference on Learning Representations}, 2018.

\bibitem[Luo et~al.(2024)Luo, Hou, Li, Chang, Liu, Wang, and Shan]{luo2024m}
Mingshuang Luo, Ruibing Hou, Zhuo Li, Hong Chang, Zimo Liu, Yaowei Wang, and Shiguang Shan.
\newblock M\textsuperscript{3} gpt: An advanced multimodal, multitask framework for motion comprehension and generation.
\newblock \emph{Advances in Neural Information Processing Systems}, 37:\penalty0 28051--28077, 2024.

\bibitem[McLean et~al.(2025)McLean, Meendering, Swartz, Gabbay, Olsen, Jacobs, Rosen, de~Bree, Garcia, Merrill, Sandakly, Buffalini, Jain, Krenn, Kumar, Markovic, Ng, Prada, Saba, Zhang, Agrawal, Godisart, Richard, and Zollhoefer]{embody3d}
Claire McLean, Makenzie Meendering, Tristan Swartz, Orri Gabbay, Alexandra Olsen, Rachel Jacobs, Nicholas Rosen, Philippe de Bree, Tony Garcia, Gadsden Merrill, Jake Sandakly, Julia Buffalini, Neham Jain, Steven Krenn, Moneish Kumar, Dejan Markovic, Evonne Ng, Fabian Prada, Andrew Saba, Siwei Zhang, Vasu Agrawal, Tim Godisart, Alexander Richard, and Michael Zollhoefer.
\newblock Embody 3d: A large-scale multimodal motion and behavior dataset, 2025.

\bibitem[Morris et~al.(2004)Morris, Maier, and Green]{morris2004}
Andrew~Cameron Morris, Viktoria Maier, and Phil~D Green.
\newblock From wer and ril to mer and wil: improved evaluation measures for connected speech recognition.
\newblock In \emph{Interspeech}, pages 2765--2768, 2004.

\bibitem[Mughal et~al.(2024)Mughal, Dabral, Habibie, Donatelli, Habermann, and Theobalt]{mughal2024convofusion}
Muhammad~Hamza Mughal, Rishabh Dabral, Ikhsanul Habibie, Lucia Donatelli, Marc Habermann, and Christian Theobalt.
\newblock Convofusion: Multi-modal conversational diffusion for co-speech gesture synthesis.
\newblock In \emph{CVPR}, pages 1388--1398, 2024.

\bibitem[Ng et~al.(2022)Ng, Joo, Hu, Li, Darrell, Kanazawa, and Ginosar]{Ng_2022_CVPR}
Evonne Ng, Hanbyul Joo, Liwen Hu, Hao Li, Trevor Darrell, Angjoo Kanazawa, and Shiry Ginosar.
\newblock Learning to listen: Modeling non-deterministic dyadic facial motion.
\newblock In \emph{CVPR}, pages 20395--20405, 2022.

\bibitem[Ng et~al.(2023{\natexlab{a}})Ng, Liu, and Kennedy]{ng2023takes}
Eley Ng, Ziang Liu, and Monroe Kennedy.
\newblock It takes two: Learning to plan for human-robot cooperative carrying.
\newblock In \emph{2023 IEEE International Conference on Robotics and Automation (ICRA)}, pages 7526--7532. IEEE, 2023{\natexlab{a}}.

\bibitem[Ng et~al.(2023{\natexlab{b}})Ng, Subramanian, Klein, Kanazawa, Darrell, and Ginosar]{ng2023text2listen}
Evonne Ng, Sanjay Subramanian, Dan Klein, Angjoo Kanazawa, Trevor Darrell, and Shiry Ginosar.
\newblock Can language models learn to listen?
\newblock In \emph{ICCV}, 2023{\natexlab{b}}.

\bibitem[Ng et~al.(2024)Ng, Romero, Bagautdinov, Bai, Darrell, Kanazawa, and Richard]{ng2024audio}
Evonne Ng, Javier Romero, Timur Bagautdinov, Shaojie Bai, Trevor Darrell, Angjoo Kanazawa, and Alexander Richard.
\newblock From audio to photoreal embodiment: Synthesizing humans in conversations.
\newblock In \emph{CVPR}, pages 1001--1010, 2024.

\bibitem[Northcutt et~al.(2020)Northcutt, Zha, Lovegrove, and Newcombe]{northcutt2020egocom}
Curtis~G Northcutt, Shengxin Zha, Steven Lovegrove, and Richard Newcombe.
\newblock Egocom: A multi-person multi-modal egocentric communications dataset.
\newblock \emph{IEEE Transactions on Pattern Analysis and Machine Intelligence}, 45\penalty0 (6):\penalty0 6783--6793, 2020.

\bibitem[Oertel et~al.(2020)Oertel, Castellano, Chetouani, Nasir, Obaid, Pelachaud, and Peters]{groupengagehri2020}
Catharine Oertel, Ginevra Castellano, Mohamed Chetouani, Jauwairia Nasir, Mohammad Obaid, Catherine Pelachaud, and Christopher Peters.
\newblock Engagement in human-agent interaction: An overview.
\newblock \emph{Frontiers in Robotics and AI}, Volume 7 - 2020, 2020.

\bibitem[{OpenAI}(2025)]{chatgpt}
{OpenAI}.
\newblock {ChatGPT (May 6 version)}.
\newblock \url{https://chat.openai.com/}, 2025.

\bibitem[Park et~al.(2025)Park, Choi, and Yun]{park2025unified}
Jeongeun Park, Sungjoon Choi, and Sangdoo Yun.
\newblock A unified framework for motion reasoning and generation in human interaction.
\newblock In \emph{Proceedings of the IEEE/CVF International Conference on Computer Vision}, pages 10698--10707, 2025.

\bibitem[Peng et~al.(2023)Peng, Mao, and Wu]{peng2023trajectory}
Xiaogang Peng, Siyuan Mao, and Zizhao Wu.
\newblock Trajectory-aware body interaction transformer for multi-person pose forecasting.
\newblock In \emph{Proceedings of the IEEE/CVF conference on computer vision and pattern recognition}, pages 17121--17130, 2023.

\bibitem[Radford et~al.(2023)Radford, Kim, Xu, Brockman, McLeavey, and Sutskever]{radford2023robust}
Alec Radford, Jong~Wook Kim, Tao Xu, Greg Brockman, Christine McLeavey, and Ilya Sutskever.
\newblock Robust speech recognition via large-scale weak supervision.
\newblock In \emph{International Conference on Machine Learning}, pages 28492--28518. PMLR, 2023.

\bibitem[Raman et~al.(2022)Raman, Vargas~Quiros, Tan, Islam, Gedik, and Hung]{raman2022conflab}
Chirag Raman, Jose Vargas~Quiros, Stephanie Tan, Ashraful Islam, Ekin Gedik, and Hayley Hung.
\newblock Conflab: A data collection concept, dataset, and benchmark for machine analysis of free-standing social interactions in the wild.
\newblock \emph{Advances in Neural Information Processing Systems}, 35:\penalty0 23701--23715, 2022.

\bibitem[Rasenberg et~al.(2022)Rasenberg, Pouw, Özyürek, and Dingemanse]{multimodal_comm-efficiency2022}
Marlou Rasenberg, Wim Pouw, Asli Özyürek, and Mark Dingemanse.
\newblock The multimodal nature of communicative efficiency in social interaction.
\newblock \emph{Scientific Reports}, 12, 2022.

\bibitem[Rempe et~al.(2020)Rempe, Guibas, Hertzmann, Russell, Villegas, and Yang]{rempe2020contact}
Davis Rempe, Leonidas~J Guibas, Aaron Hertzmann, Bryan Russell, Ruben Villegas, and Jimei Yang.
\newblock Contact and human dynamics from monocular video.
\newblock In \emph{Computer Vision--ECCV 2020: 16th European Conference, Glasgow, UK, August 23--28, 2020, Proceedings, Part V 16}, pages 71--87. Springer, 2020.

\bibitem[Ribeiro-Gomes et~al.(2024)Ribeiro-Gomes, Cai, Milacski, Wu, Prakash, Takagi, Aubel, Kim, Bernardino, and De~La~Torre]{ribeiro2024motiongpt}
Jose Ribeiro-Gomes, Tianhui Cai, Zolt{\'a}n~A Milacski, Chen Wu, Aayush Prakash, Shingo Takagi, Amaury Aubel, Daeil Kim, Alexandre Bernardino, and Fernando De~La~Torre.
\newblock Motiongpt: Human motion synthesis with improved diversity and realism via gpt-3 prompting.
\newblock In \emph{Proceedings of the IEEE/CVF Winter Conference on Applications of Computer Vision}, pages 5070--5080, 2024.

\bibitem[Russell and Harte(2025)]{russell2025visual}
Sam~O’Connor Russell and Naomi Harte.
\newblock Visual cues enhance predictive turn-taking for two-party human interaction.
\newblock In \emph{Findings of the Association for Computational Linguistics: ACL 2025}, pages 209--221, 2025.

\bibitem[Siniukov et~al.(2025)Siniukov, Chang, Tran, Gong, Chaubey, and Soleymani]{siniukov2025ditailistener}
Maksim Siniukov, Di Chang, Minh Tran, Hongkun Gong, Ashutosh Chaubey, and Mohammad Soleymani.
\newblock Ditailistener: Controllable high fidelity listener video generation with diffusion.
\newblock In \emph{Proceedings of the IEEE/CVF International Conference on Computer Vision}, pages 11991--12001, 2025.

\bibitem[Siyao et~al.(2024)Siyao, Gu, Yang, Lin, Liu, Ding, Yang, and Loy]{siyao2024duolando}
Li Siyao, Tianpei Gu, Zhitao Yang, Zhengyu Lin, Ziwei Liu, Henghui Ding, Lei Yang, and Chen~Change Loy.
\newblock Duolando: Follower gpt with off-policy reinforcement learning for dance accompaniment.
\newblock In \emph{ICLR}, 2024.

\bibitem[Spitale et~al.(2024)Spitale, Parreira, Stiber, Axelsson, Kara, Kankariya, Huang, Jung, Ju, and Gunes]{multimodal_roboterror2024}
Micol Spitale, Maria~Teresa Parreira, Maia Stiber, Minja Axelsson, Neval Kara, Garima Kankariya, Chien-Ming Huang, Malte Jung, Wendy Ju, and Hatice Gunes.
\newblock Err@hri 2024 challenge: Multimodal detection of errors and failures in human-robot interactions.
\newblock In \emph{Proceedings of the 26th International Conference on Multimodal Interaction}, page 652–656, New York, NY, USA, 2024. Association for Computing Machinery.

\bibitem[stable-ts()]{stable-ts}
stable-ts.
\newblock stable-ts.
\newblock \url{https://github.com/jianfch/stable-ts}.
\newblock Accessed: 2025-05-15.

\bibitem[Tanke et~al.(2023{\natexlab{a}})Tanke, Kwon, Mueller, Doering, and Gall]{tanke2023humans}
Julian Tanke, Oh-Hun Kwon, Felix~B Mueller, Andreas Doering, and Juergen Gall.
\newblock Humans in kitchens: a dataset for multi-person human motion forecasting with scene context.
\newblock \emph{Advances in Neural Information Processing Systems}, 36:\penalty0 10184--10196, 2023{\natexlab{a}}.

\bibitem[Tanke et~al.(2023{\natexlab{b}})Tanke, Zhang, Zhao, Tang, Cai, Wang, Wu, Gall, and Keskin]{tanke2023social}
Julian Tanke, Linguang Zhang, Amy Zhao, Chengcheng Tang, Yujun Cai, Lezi Wang, Po-Chen Wu, Juergen Gall, and Cem Keskin.
\newblock Social diffusion: Long-term multiple human motion anticipation.
\newblock In \emph{ICCV}, pages 9601--9611, 2023{\natexlab{b}}.

\bibitem[Trabelsi et~al.(2019)Trabelsi, Varadarajan, Zhang, Jabri, Pei, Smach, Bouallegue, and Moulin]{multimodal_socialdynamics2019}
Rim Trabelsi, Jagannadan Varadarajan, Le Zhang, Issam Jabri, Yong Pei, Fethi Smach, Ammar Bouallegue, and Pierre Moulin.
\newblock Understanding the dynamics of social interactions: A multi-modal multi-view approach.
\newblock \emph{ACM Trans. Multimedia Comput. Commun. Appl.}, 15\penalty0 (1s), 2019.

\bibitem[Umair et~al.(2024)Umair, Sarathy, and de~Ruiter]{umair2024large}
Muhammad Umair, Vasanth Sarathy, and JP de Ruiter.
\newblock Large language models know what to say but not when to speak.
\newblock \emph{arXiv preprint arXiv:2410.16044}, 2024.

\bibitem[Urakami and Seaborn(2023)]{noC4hri2023}
Jacqueline Urakami and Katie Seaborn.
\newblock Nonverbal cues in human–robot interaction: A communication studies perspective.
\newblock \emph{J. Hum.-Robot Interact.}, 12\penalty0 (2), 2023.

\bibitem[Vendrow et~al.(2023)Vendrow, Le, Cai, and Rezatofighi]{vendrow2023jrdb}
Edward Vendrow, Duy~Tho Le, Jianfei Cai, and Hamid Rezatofighi.
\newblock Jrdb-pose: A large-scale dataset for multi-person pose estimation and tracking.
\newblock In \emph{Proceedings of the IEEE/CVF Conference on Computer Vision and Pattern Recognition}, pages 4811--4820, 2023.

\bibitem[Vogel et~al.(2018)Vogel, Meyer, and Harendza]{noc4empathy2018}
Daniela Vogel, Marco Meyer, and Sigrid Harendza.
\newblock Verbal and non-verbal communication skills including empathy during history taking of undergraduate medical students.
\newblock \emph{BMC Medical Education}, 18, 2018.

\bibitem[Voultsiou and Moussiades(2025)]{voultsiou2025systematic}
Evdokia Voultsiou and Lefteris Moussiades.
\newblock A systematic review of ai, vr, and llm applications in special education: Opportunities, challenges, and future directions.
\newblock \emph{Education and Information Technologies}, pages 1--41, 2025.

\bibitem[Vrij et~al.(2019)Vrij, Hartwig, and Granhag]{noc4deception2019}
Aldert Vrij, Maria Hartwig, and Pär~Anders Granhag.
\newblock Reading lies: Nonverbal communication and deception.
\newblock \emph{Annual review of psychology}, 70:\penalty0 295—317, 2019.

\bibitem[Wang et~al.(2024)Wang, Chen, Jia, Li, Zhang, Zhang, Liu, Zhu, Liang, and Huang]{Wang_2024_CVPR}
Zan Wang, Yixin Chen, Baoxiong Jia, Puhao Li, Jinlu Zhang, Jingze Zhang, Tengyu Liu, Yixin Zhu, Wei Liang, and Siyuan Huang.
\newblock Move as you say interact as you can: Language-guided human motion generation with scene affordance.
\newblock In \emph{Proceedings of the IEEE/CVF Conference on Computer Vision and Pattern Recognition (CVPR)}, pages 433--444, 2024.

\bibitem[Wiemann and Knapp(2017)]{wiemann2017turn}
John~M Wiemann and Mark~L Knapp.
\newblock Turn-taking in conversations.
\newblock \emph{Communication theory}, pages 226--245, 2017.

\bibitem[Woodard and Nelson(1982)]{woodard1982}
J.P. Woodard and J.T. Nelson.
\newblock An information theoretic measure of speech recognition performance.
\newblock \emph{Workshop on standardisation for speech I/O technology}, 1982.

\bibitem[Wu et~al.(2024)Wu, Fei, Qu, Ji, and Chua]{wu2024next}
Shengqiong Wu, Hao Fei, Leigang Qu, Wei Ji, and Tat-Seng Chua.
\newblock Next-gpt: Any-to-any multimodal llm.
\newblock In \emph{Forty-first International Conference on Machine Learning}, 2024.

\bibitem[Xiao et~al.(2024)Xiao, Xie, Xu, Chen, and Zhang]{xiao2024multi}
Peng Xiao, Yi Xie, Xuemiao Xu, Weihong Chen, and Huaidong Zhang.
\newblock Multi-person pose forecasting with individual interaction perceptron and prior learning.
\newblock In \emph{European Conference on Computer Vision}, pages 402--419. Springer, 2024.

\bibitem[Xu and LC(2022)]{groupcohesionhri2022}
Hongshen Xu and Ray LC.
\newblock Cohesiveness of robots in groups affects the perception of social rejection by human observers.
\newblock In \emph{2022 17th ACM/IEEE International Conference on Human-Robot Interaction (HRI)}, pages 1100--1104, 2022.

\bibitem[Xu et~al.(2024)Xu, Zhou, Yan, Jin, Zhu, Rao, Yang, and Zeng]{xu2024regennet}
Liang Xu, Yizhou Zhou, Yichao Yan, Xin Jin, Wenhan Zhu, Fengyun Rao, Xiaokang Yang, and Wenjun Zeng.
\newblock Regennet: Towards human action-reaction synthesis.
\newblock In \emph{Proceedings of the IEEE/CVF Conference on Computer Vision and Pattern Recognition}, pages 1759--1769, 2024.

\bibitem[Yi et~al.(2023)Yi, Liang, Liu, Cao, Wen, Bolkart, Tao, and Black]{yi2022generating}
Hongwei Yi, Hualin Liang, Yifei Liu, Qiong Cao, Yandong Wen, Timo Bolkart, Dacheng Tao, and Michael~J Black.
\newblock Generating holistic 3d human motion from speech.
\newblock In \emph{CVPR}, 2023.

\bibitem[Yu et~al.(2025)Yu, Zhang, Chen, Xiang, Fang, Niebles, and Adeli]{yu2025socialgen}
Heng Yu, Juze Zhang, Changan Chen, Tiange Xiang, Yusu Fang, Juan~Carlos Niebles, and Ehsan Adeli.
\newblock Socialgen: Modeling multi-human social interaction with language models.
\newblock \emph{arXiv preprint arXiv:2503.22906}, 2025.

\bibitem[{\.Z}arkowski(2019)]{zarkowski2019multi}
Mateusz {\.Z}arkowski.
\newblock Multi-party turn-taking in repeated human--robot interactions: an interdisciplinary evaluation.
\newblock \emph{International Journal of Social Robotics}, 11\penalty0 (5):\penalty0 693--707, 2019.

\bibitem[Zhan et~al.(2024)Zhan, Dai, Ye, Zhou, Zhang, Liu, Zhang, Yuan, Zhang, Li, et~al.]{zhan2024anygpt}
Jun Zhan, Junqi Dai, Jiasheng Ye, Yunhua Zhou, Dong Zhang, Zhigeng Liu, Xin Zhang, Ruibin Yuan, Ge Zhang, Linyang Li, et~al.
\newblock Anygpt: Unified multimodal llm with discrete sequence modeling.
\newblock \emph{arXiv preprint arXiv:2402.12226}, 2024.

\bibitem[Zhang et~al.(2023{\natexlab{a}})Zhang, Zhang, Cun, Huang, Zhang, Zhao, Lu, and Shen]{zhang2023generating}
Jianrong Zhang, Yangsong Zhang, Xiaodong Cun, Shaoli Huang, Yong Zhang, Hongwei Zhao, Hongtao Lu, and Xi Shen.
\newblock T2m-gpt: Generating human motion from textual descriptions with discrete representations.
\newblock In \emph{CVPR}, 2023{\natexlab{a}}.

\bibitem[Zhang et~al.(2024{\natexlab{a}})Zhang, Zhang, Song, Shi, Zhao, Shi, Yu, Xu, and Wang]{zhang2024hoi}
Juze Zhang, Jingyan Zhang, Zining Song, Zhanhe Shi, Chengfeng Zhao, Ye Shi, Jingyi Yu, Lan Xu, and Jingya Wang.
\newblock Hoi-m3: Capture multiple humans and objects interaction within contextual environment.
\newblock In \emph{Proceedings of the IEEE/CVF Conference on Computer Vision and Pattern Recognition}, pages 516--526, 2024{\natexlab{a}}.

\bibitem[Zhang et~al.(2024{\natexlab{b}})Zhang, Cai, Pan, Hong, Guo, Yang, and Liu]{zhang2024motiondiffuse}
Mingyuan Zhang, Zhongang Cai, Liang Pan, Fangzhou Hong, Xinying Guo, Lei Yang, and Ziwei Liu.
\newblock Motiondiffuse: Text-driven human motion generation with diffusion model.
\newblock \emph{IEEE transactions on pattern analysis and machine intelligence}, 46\penalty0 (6):\penalty0 4115--4128, 2024{\natexlab{b}}.

\bibitem[Zhang et~al.(2024{\natexlab{c}})Zhang, Jin, Gu, Hong, Cai, Huang, Zhang, Guo, Yang, He, et~al.]{zhang2024large}
Mingyuan Zhang, Daisheng Jin, Chenyang Gu, Fangzhou Hong, Zhongang Cai, Jingfang Huang, Chongzhi Zhang, Xinying Guo, Lei Yang, Ying He, et~al.
\newblock Large motion model for unified multi-modal motion generation.
\newblock In \emph{European Conference on Computer Vision}, pages 397--421. Springer, 2024{\natexlab{c}}.

\bibitem[Zhang* et~al.(2020)Zhang*, Kishore*, Wu*, Weinberger, and Artzi]{bert-score}
Tianyi Zhang*, Varsha Kishore*, Felix Wu*, Kilian~Q. Weinberger, and Yoav Artzi.
\newblock Bertscore: Evaluating text generation with bert.
\newblock In \emph{International Conference on Learning Representations}, 2020.

\bibitem[Zhang et~al.(2023{\natexlab{b}})Zhang, Zhang, Li, Zhou, and Qiu]{zhang2023speechtokenizer}
Xin Zhang, Dong Zhang, Shimin Li, Yaqian Zhou, and Xipeng Qiu.
\newblock Speechtokenizer: Unified speech tokenizer for speech large language models.
\newblock \emph{arXiv preprint arXiv:2308.16692}, 2023{\natexlab{b}}.

\bibitem[Zhang et~al.(2021)Zhang, Li, An, Li, Yu, and Liu]{lightcap2021}
Yuxiang Zhang, Zhe Li, Liang An, Mengcheng Li, Tao Yu, and Yebin Liu.
\newblock Light-weight multi-person total capture using sparse multi-view cameras.
\newblock In \emph{IEEE International Conference on Computer Vision}, 2021.

\bibitem[Zhou et~al.(2019)Zhou, Barnes, Lu, Yang, and Li]{zhou2019continuity}
Yi Zhou, Connelly Barnes, Jingwan Lu, Jimei Yang, and Hao Li.
\newblock On the continuity of rotation representations in neural networks.
\newblock In \emph{Proceedings of the IEEE/CVF conference on computer vision and pattern recognition}, pages 5745--5753, 2019.

\bibitem[Zhou et~al.(2024)Zhou, Wan, and Wang]{zhou2024avatargpt}
Zixiang Zhou, Yu Wan, and Baoyuan Wang.
\newblock Avatargpt: All-in-one framework for motion understanding planning generation and beyond.
\newblock In \emph{Proceedings of the IEEE/CVF Conference on Computer Vision and Pattern Recognition}, pages 1357--1366, 2024.

\bibitem[Zhu et~al.(2023)Zhu, Lin, Ning, Yan, Cui, Wang, Pang, Jiang, Zhang, Li, et~al.]{zhu2023languagebind}
Bin Zhu, Bin Lin, Munan Ning, Yang Yan, Jiaxi Cui, HongFa Wang, Yatian Pang, Wenhao Jiang, Junwu Zhang, Zongwei Li, et~al.
\newblock Languagebind: Extending video-language pretraining to n-modality by language-based semantic alignment.
\newblock \emph{arXiv preprint arXiv:2310.01852}, 2023.

\bibitem[Zhu et~al.(2025)Zhu, Zhang, Rong, Hu, Liang, and Ge]{zhu2025infp}
Yongming Zhu, Longhao Zhang, Zhengkun Rong, Tianshu Hu, Shuang Liang, and Zhipeng Ge.
\newblock Infp: Audio-driven interactive head generation in dyadic conversations.
\newblock In \emph{Proceedings of the IEEE/CVF Conference on Computer Vision and Pattern Recognition}, pages 10667--10677, 2025.

\end{thebibliography}
}
\clearpage
\setcounter{page}{1}
\maketitlesupplementary
\appendix
\crefalias{section}{appendix}

\section{Data Pre-processing}
\label{app:data}
\subsection{Transcription}
\begin{figure}[t]
  \centering
  \includegraphics[width=\linewidth]{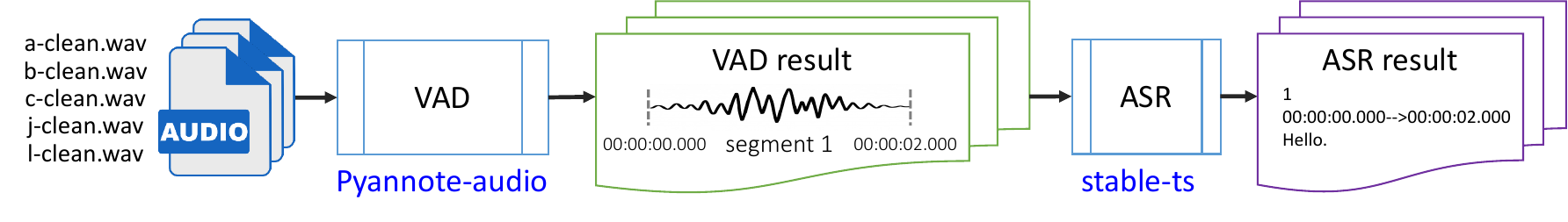}
   \caption{Transcription workflow. Utterances are extracted by pyannote-audio \cite{bredin2021end}, and then transcribed using stable-ts \cite{stable-ts}.
   }
  \label{fig:transcript_flow}
\end{figure}

As illustrated in \cref{fig:transcript_flow}, to segment utterances from the per-participant audio in the DnD Gesture Dataset \cite{mughal2024convofusion}, we first apply Voice Activity Detection (VAD) using Pyannote-audio \cite{bredin2021end}. We then perform Automatic Speech Recognition (ASR) using stable-ts \cite{stable-ts} to obtain the transcription and corresponding timestamps for each utterance. 

\subsection{Data Chunking}
For each utterance, we extract a 64-frame segment (approximately 2.56 seconds) starting at the utterance onset. Segments from the target participant's utterances are assigned to the speaking subset, while all others are categorized as the listening subset.
For dataset balance, the listening subset is downsampled to match the size of the speaking subset.

\begin{figure*}[t]
  \centering
  \includegraphics[width=\textwidth]{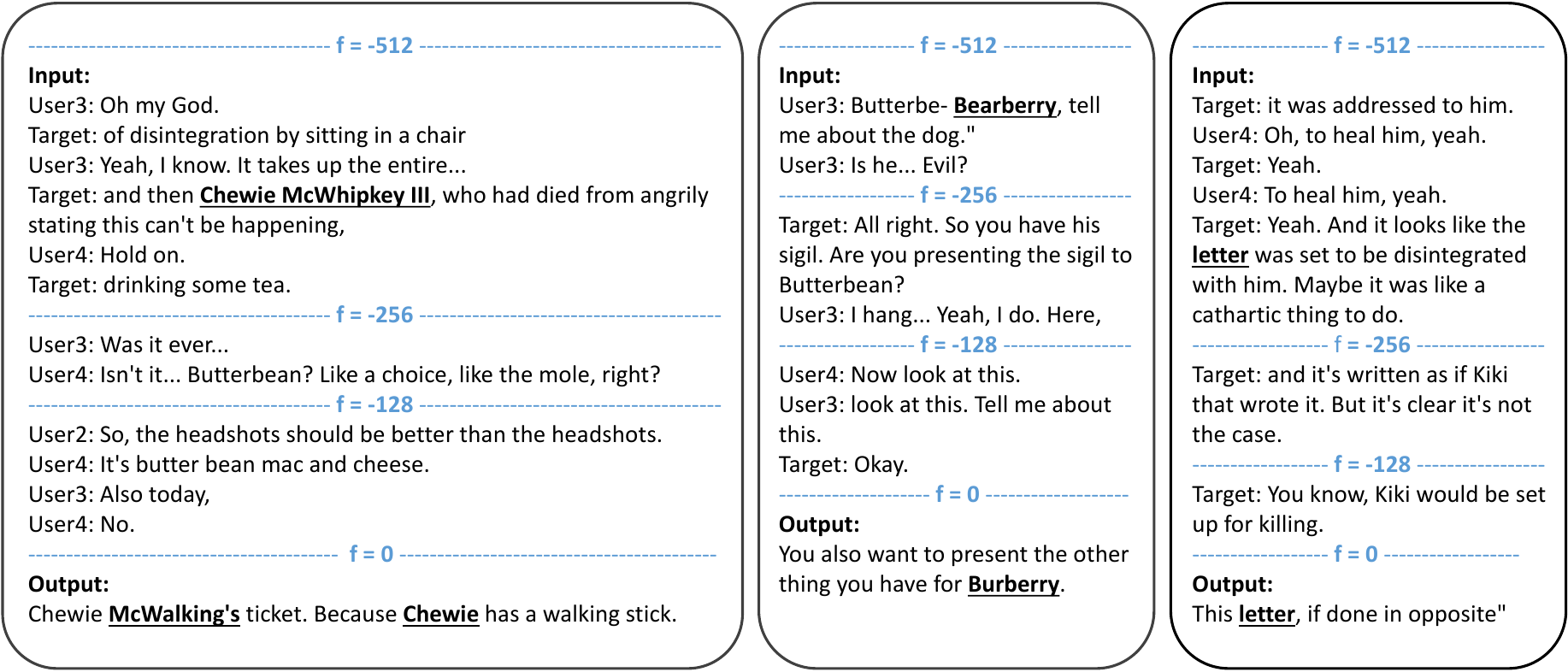}
   \caption{Three examples of the conversational chunks extracted from the DnD Gesture Dataset \cite{mughal2024convofusion}. Keywords essential for generating in-context responses are underlined. $f$ indicates the frame index. The output speaker in each example is the target participant.
  }
  \label{fig:history_example}
\end{figure*}

To make a chunk, each segment is paired with a speech history up to 512 frames (approximately 20.48 seconds) preceding the segment to support modeling long-form, engaging conversations. We select 512 frames because this window provides sufficient contextual information for generating coherent and contextually appropriate responses. Some examples can be found in \cref{fig:history_example}.
Unlike linguistic context, the motion observations include only 64 frames (approximately 2.56 seconds in duration) for all participants.
In total, the processed dataset contains 8,926 chunks for training and validation, and 3,028 chunks for testing.

\subsection{Speech Style Extraction}
We first extract the utterance audio using the timestamps obtained from ASR, and then utilize StyleTTS 2 \cite{li2023styletts} to extract the corresponding speech style.

\subsection{Motion Pre-processing}
The captured motion data are originally in BVH format, containing a global translation and 54 joint rotations per frame. We convert the motion data into a 327-dimensional representation. The first three dimensions correspond to global translation, while the remaining 324 dimensions represent the 6D rotations \cite{zhou2019continuity} of the 54 body and hand joints.

\section{Training Details}
\label{app:train}

\subsection{Loss Function}
The full loss function is defined as:
\begin{equation}
\mathcal{L}_{\text{total}} = \lambda_{\text{text}} \cdot \mathcal{L}_{\text{text}} + \lambda_{\text{style}} \cdot \mathcal{L}_{\text{style}} + \lambda_{\text{state}} \cdot \mathcal{L}_{\text{state}} + \mathcal{L}_{\text{motion}},
\end{equation}
where $\mathcal{L}_{\text{text}}$, $\mathcal{L}_{\text{style}}$, $\mathcal{L}_{\text{state}}$, and $\mathcal{L}_{\text{motion}}$ represent the losses associated with text token, speech style feature, speaking-state score, and motion, respectively. The coefficients $\lambda_{\text{text}}$, $\lambda_{\text{style}}$, and $\lambda_{\text{state}}$ are the respective weighting factors.
The motion loss $\mathcal{L}_{\text{motion}}$ is defined as:
\begin{equation}
\begin{aligned}
\mathcal{L}_{\text{motion}} = \lambda_{\text{repr}} \cdot \mathcal{L}_{\text{repr}} &+ \lambda_{\text{keypoint}} \cdot \mathcal{L}_{\text{keypoint}} \\ &+ \lambda_{\text{root}} \cdot \mathcal{L}_{\text{root}} + \lambda_{\text{reg}} \cdot \mathcal{L}_{\text{reg}},
\end{aligned}
\end{equation}
where $\mathcal{L}_{\text{repr}}$, $\mathcal{L}_{\text{root}}$, and $\mathcal{L}_{\text{keypoint}}$ are L2 losses applied to the motion representation, the global translation, and the localized joint positions, respectively. The corresponding weighting coefficients are $\lambda_{\text{repr}}$, $\lambda_{\text{keypoint}}$, $\lambda_{\text{root}}$, and $\lambda_{\text{reg}}$. The regularization term, $\mathcal{L}_{\text{reg}}$, penalizes unnatural motion characteristics, including unrealistic joint velocities, excessive accelerations, and implausible foot-ground contact patterns.

The loss weights for the training objectives are set as follows: $\lambda_{\text{text}} = 1.0$, $\lambda_{\text{style}} = 100.0$, $\lambda_{\text{state}} = 0.4$, $\lambda_{\text{repr}} = 0.5$, $\lambda_{\text{keypoint}} = 1000.0$, $\lambda_{\text{root}} = 50$, and $\lambda_{\text{reg}} = 10$.

\subsection{Multimodal Instruction Template}
We adapt the instruction template to incorporate speech style, full-body motion, and social cues for polyadic interaction. An example template is shown in \cref{fig:template}.

\myheading{Adaptation for Multimodal Interaction} 
To support multimodal interaction, we extend the instruction template by assigning distinct roles to each participant and introducing additional special tokens to represent motion observations and social cues. 
    
\myheading{Adaptation for Multimodal Outputs} 
To accommodate variable-length text output while maintaining a fixed-length 64-frame motion sequence, text decoding is terminated either upon the generation of the first end-of-turn token, \texttt{<eot\_id>}, or upon reaching a maximum of 64 generated words, whichever occurs first. Following this token, one additional embedding is decoded as the speech style, and the subsequent 64 embeddings are decoded as motion.

\begin{figure}
\begin{adjustbox}{width=\linewidth,center}
\begin{tcolorbox}[
  colback=gray!10,
  width=0.6\textwidth, 
  title=Interaction Template,
  coltitle=white,
  colbacktitle=gray!60,
  sharp corners=south,
  boxrule=0.8pt,
  fonttitle=\bfseries, 
  fontupper=\small, 
  ]
\textbf{Input:}\\
User1: \texttt{<TEXT>...<TEXT><eot\_id><STYLE>}\\
User3: \texttt{<TEXT>...<TEXT><eot\_id><STYLE>}\\
Target: \texttt{<TEXT>...<TEXT><eot\_id><STYLE>}\\
\texttt{...}\\
Motion: \texttt{<POSE>...<POSE>}\\
Social: \texttt{<SCUE><SCUE>}\\
Target: \\
\textbf{Output:}\\
\texttt{<TEXT>...<TEXT><eot\_id><STYLE><POSE>...<POSE>}
\end{tcolorbox}
\end{adjustbox}
\caption{Multimodal instruction template for polyadic interaction. \texttt{<TEXT>} denotes text embeddings, \texttt{<STYLE>} represents speech style embeddings, \texttt{<POSE>} corresponds to motion embeddings, and \texttt{<SCUE>} refers to social cue embeddings.}
\label{fig:template}
\end{figure}

\section{Evaluation Details}
\label{app:eval}

\subsection{Speech Metrics}
\myheading{SIM} 
To assess voice similarity, we follow a controlled evaluation approach to isolate the effect of the generated text. Specifically, we synthesize audio using the ground-truth text and the generated speech style. This ensures comparability and avoids cases where no speech is generated. Voice similarity is evaluated only for the speaking subset.

\myheading{BeatAlignDiff} 
We adopt the beat alignment metric as defined in \cite{mughal2024convofusion}, with the difference computed as:
\begin{equation}
\text{BeatAlignDiff} = |\text{BeatAlign}_{\text{pred}} - \text{BeatAlign}_{\text{GT}}|.
\end{equation}
Audio beats are extracted from onset timestamps of synthesized speech, while motion beats are identified as local minima in the velocity magnitudes of selected body joints.

Beat alignment is computed only when both ground-truth and generated speech are available. 
To ensure that the evaluation does not benefit from skipping chunks with missing generated speech, we analyzed one evaluation run and found that 28 out of 1,514 speaking chunks (0.84\%) generated by SOLAMI \cite{jiang2024solami} lacked generated speech. In contrast, our method produces only 3 such cases (0.19\%).

\begin{table*}[t]
\caption{Comparison of \methodname with baselines with standard deviation over five runs. \textsuperscript{\dag}: Infer speaking state based on the generated text.\textsuperscript{*}: Synthesize speech using prompts from test set audio. Diversity closer to the ground truth (117.09) indicates better performance.
}
\centering
\begin{adjustbox}{width=\textwidth,center}
\begin{tabular}{lccccccccc}
\toprule
\multirow{3}[3]{*}{Method} & \multicolumn{5}{c}{Motion}  & \multicolumn{3}{c}{Speech} & \multicolumn{1}{c}{State} \\ \cmidrule(lr){2-6} \cmidrule(lr){7-9} \cmidrule(lr){10-10}
                           & Root$\downarrow$   & MPJPE$\downarrow$ & \multirow{2}{*}{FID$\downarrow$} & \multirow{2}{*}{Diversity$\rightarrow$} & BeatAlign & BERT  & \multirow{2}{*}{WER$\downarrow$} & \multirow{2}{*}{SIM$\uparrow$}  &  \multirow{2}{*}{AP$\uparrow$} \\ 
                           & (mm)   & (mm) &   &  & Diff.$\downarrow$ & Score$\uparrow$ & &  & \\

\midrule
Random                     & $140.4^{\pm 0.90}$  & $200.4^{\pm 0.90}$ & $17.82^{\pm 0.035}$  & $120.46^{\pm 0.406}$ & $0.018^{\pm 0.001}$  & $0.458^{\pm 0.0019}$  & $1.699^{\pm 0.014}$  & $0.494^{\pm 0.0048}$   & $0.50^{\pm 0.003}$\textsuperscript{\dag} \\
NN cond.                   & $134.7^{\pm 0.00}$  & $187.7^{\pm 0.00}$ & $16.36^{\pm 0.000}$  & $105.42^{\pm 0.000}$ & $0.023^{\pm 0.000}$ & $0.451^{\pm 0.0000}$  & $2.075^{\pm 0.000}$  & $0.520^{\pm 0.0000}$   & $0.52^{\pm 0.000}$\textsuperscript{\dag} \\
\midrule
LLM + ConvoFusion \cite{mughal2024convofusion}   & $125.7^{\pm 0.19}$ & $170.1^{\pm 0.06}$ & $18.57^{\pm 0.012}$ & $72.45^{\pm 0.139}$  &  -     & $0.388^{\pm  0.0007}$ & $13.318^{\pm 0.073}$  &  - & $0.50^{\pm 0.000}$\textsuperscript{\dag} \\
LM-L2L Adapted \cite{ng2023text2listen}   & $185.2^{\pm 0.00}$  & $187.6^{\pm 0.00}$ & $17.22^{\pm 0.000}$  & $\textbf{116.36}^{\pm 0.000}$ & -  & -  & -  & -  & -\\
SOLAMI \cite{jiang2024solami}             & $188.6^{\pm 1.22}$  & $180.9^{\pm 2.08}$ & $14.86^{\pm 0.122}$ & $100.13^{\pm 1.575}$ & $0.061^{\pm 0.005}$ & $0.428^{\pm 0.0002}$ & $1.854^{\pm 0.015}$ & {$0.745^{\pm 0.0001}$}\textsuperscript{*} & $0.50^{\pm 0.001}$\textsuperscript{\dag}\\
\midrule
Ours  & $\textbf{108.7}^{\pm 0.16}$ & $\textbf{144.9}^{\pm 0.07}$ & $\textbf{12.18}^{\pm 0.007}$ & $113.32^{\pm 0.055}$ & $\textbf{0.007}^{\pm 0.001}$ & $\textbf{0.508}^{\pm 0.0011}$  & $\textbf{1.436}^{\pm 0.008}$ & $\textbf{0.642}^{\pm  0.0010}$  & $\textbf{0.67}^{\pm 0.000}$\\
\bottomrule
\end{tabular}
\end{adjustbox}
\label{tab:compare_baseline_std}
\end{table*}

\subsection{Motion Metric}
\label{sub:motion_metric}
\myheading{Motion Evaluator}
A motion evaluator is pre-trained for both FID and Diversity, since both are calculated in a latent feature space. 
Following prior motion generation works \cite{guo2022generating, ng2023text2listen, ribeiro2024motiongpt, guo2022tm2t}, we adopt a convolutional encoder-decoder architecture and train it to reconstruct motions from the DnD Gesture Dataset. Hyperparameter settings are set following \cite{guo2022generating}. The resulting evaluator achieves a mean per-joint position error (MPJPE) of 20.89 mm and a root joint Euclidean distance of 21.07 mm.

\myheading{Diversity} 
We compute diversity for both the generated and ground-truth motions to compare the variability between the two distributions. Diversity quantifies the average pairwise Euclidean distance among reactive motion features within a set. Let $\mathbf{f}_1, \mathbf{f}_2, \dots, \mathbf{f}_n$ denote the feature vectors of the generated motions. The diversity is computed as:
\begin{equation}
Diversity = \frac{2}{n(n - 1)} \sum_{i=1}^{n-1} \sum_{j=i+1}^{n} \left\| \mathbf{f}_i - \mathbf{f}_j \right\|_2.
\end{equation}
We follow the DnD Group Gesture dataset and baselines \cite{ng2023text2listen, mughal2024convofusion}, measuring diversity relative to ground-truth statistics, where values closer to ground-truth indicate variability better matching real interactions.

\subsection{Social Semantics Metrics}
We use Mean Angular Error $\text{MAE}_\text{head}$ and social cue score error to evaluate the social semantics of the generated reaction. $\text{MAE}_\text{head}$ is the rotation angle between the generated and ground-truth head orientations. The social cue score error semantically represents the difference in the attention that each non-target participant receives from the target participant. It is computed as the difference between the non-target participant's social cue scores derived from the generated and ground-truth head positions and orientations. Specifically, the social cue score for a non-target participant $i$ at frame $k$ is defined as:
\begin{equation}
s_i^k = \cos(\mathbf{u}_\text{P}^k, \mathbf{v}_{\text{P} \rightarrow i}^k),
\end{equation}
, where $\mathbf{u}_\text{P}^k$ is the head orientation of the target participant, and $\mathbf{v}_{\text{P} \rightarrow i}^k$ is the relative head vector from the target participant to the non-target participant $i$.

\subsection{Baseline Comparison with Standard Deviation }
We run all evaluations five times, and only the average results are reported in the main paper for clarity. In \cref{tab:compare_baseline_std}, we present the baseline comparisons along with standard deviations. Compared to SOLAMI, \methodname achieves not only better performance but also lower standard deviations across most metrics, indicating improved stability and consistency. The LM-L2L Adapted baseline shows no variation across runs, as its mapping from embeddings to pose space is deterministic. In contrast, the motion generated by \methodname varies based on the preceding generated text due to the autoregressive nature of LLMs.

\section{Baseline Implementation}
\label{app:baseline}

\subsection{NN Condition} 
For each chunk, we use the embeddings of the input observations to retrieve the Nearest-Neighbor (NN) from the training set, and take the corresponding response as the output. Text embeddings are obtained using the embedding layer of Llama3–8B-Instruct \cite{grattafiori2024llama}, while motion embeddings are from the motion evaluator introduced in \cref{sub:motion_metric}.

\subsection{LLM + ConvoFusion}
We utilize Llama3-8B-Instruct \cite{grattafiori2024llama} as the LLM backbone and disable the audio condition of ConvoFusion \cite{mughal2024convofusion} by setting it to unconditional tokens, following its original design. As ConvoFusion is already trained on the DnD Group Gesture dataset, finetuning is not performed for this baseline.

\begin{table*}[t]
\caption{Comparison of \methodname with further extended baselines.
SOLAMI is performed with LoRA finetuning and without pre-training. 
\textsuperscript{\dag}: Infer speaking state based on the generated text.\textsuperscript{*}: Synthesize speech with prompts from test set audio. Diversity closer to the ground truth (117.09) indicates better performance. Underline denotes the second-best results.
}
\centering
\begin{adjustbox}{width=\textwidth,center}
\begin{tabular}{lccccccccccc}
\toprule
\multirow{3}[3]{*}{Method} & \multicolumn{2}{c}{Training Data} & \multicolumn{5}{c}{Motion}  & \multicolumn{3}{c}{Speech} & \multicolumn{1}{c}{State} \\ \cmidrule(lr){2-3} \cmidrule(lr){4-8} \cmidrule(lr){9-11} \cmidrule(lr){12-12}
                        &\multirow{2}{*}{Listening} &\multirow{2}{*}{Speaking} & Root$\downarrow$   & MPJPE$\downarrow$ & \multirow{2}{*}{FID$\downarrow$} & \multirow{2}{*}{Div.$\rightarrow$} & BeatAlign & BERT  & \multirow{2}{*}{WER$\downarrow$} & \multirow{2}{*}{SIM$\uparrow$}  &  \multirow{2}{*}{AP$\uparrow$} \\ 
                         & & & (mm)   & (mm) &   &  & Diff.$\downarrow$ & Score$\uparrow$ & &  & \\

\midrule
\multirow{2}{*}{LLM + ConvoFusion \cite{mughal2024convofusion}} & \multicolumn{2}{c}{w/o finetune}    & 125.7 & \underline{170.1} & 18.57 & 72.45  &  -     & 0.388 & 13.318  &  - & 0.50\textsuperscript{\dag} \\
            & \cmark &  \cmark        & \underline{121.5}  & 170.5 & 17.43 & 67.18 &  -     & \textbf{0.511} & \textbf{1.396}  &  - & 0.56\textsuperscript{\dag} \\
\midrule
\multirow{2}{*}{LM-L2L Adapted \cite{ng2023text2listen}} & \cmark &      & 185.2  & 187.6 & 17.22  & \underline{116.36} & -  & -  & -  & -  & -\\
& \cmark &  \cmark   & 167.3  & 186.7 & 17.05  & \textbf{116.64} & -  & -  & -  & -  & -\\
\midrule
\multirow{2}{*}{SOLAMI \cite{jiang2024solami}}    &  &  \cmark           & 188.6  & 180.9 & \underline{14.86} & 100.13 & \underline{0.061} & 0.428 & 1.854 & 0.745\textsuperscript{*} & 0.50\textsuperscript{\dag}\\
 & \cmark &  \cmark                & 170.9  & 181.3 & 15.21 & 103.87 & 0.065 & 0.503 & 1.548 & 0.744\textsuperscript{*} & 0.52\textsuperscript{\dag}\\
\midrule
Ours & \cmark &  \cmark  & \textbf{108.7} & \textbf{144.9} & \textbf{12.18} & 113.32 & \textbf{0.007} & \underline{0.508}  & \underline{1.436} & \textbf{0.642}  & \textbf{0.67}\\
\bottomrule
\end{tabular}
\end{adjustbox}
\label{tab:baselines_alternative}
\end{table*}

\subsection{SOLAMI} 

\myheading{Motion Tokenizer} 
In SOLAMI \cite{jiang2024solami}, motions are first encoded into tokens. Following this approach, we train a VQ-VAE-based motion tokenizer using the network from \cite{ribeiro2024motiongpt} for DnD Gesture Dataset. Consistent SOLAMI, we decompose the pose into three groups: root position, hand rotations, and body rotations. The training loss includes L2 loss on global translation, pose representation, keypoint positions and velocities, and a commitment loss. We assign one codebook per group, each containing 512 codewords with a hidden size of 256 and a temporal depth of 2. 

The resulting tokenizer achieves an MPJPE of 90.5 mm and an Euclidean distance of 7.9 mm for the root joint position, which are comparable to the 88 mm MPJPE reported by SOLAMI on their dataset.

\myheading{Direct Training} 
Pre-training is not performed since the DnD Group Gesture dataset lacks motion captions. This also ensures a fair comparison with our \methodname.

\section{Additional Experiments}
\label{app:experiments}

\subsection{Comparison to Extended Baselines}
All baselines are originally designed for generating speaking behaviors. We further investigate how well they can handle listening reactions when provided with both speaking and listening data. Note that some adaptations require changes that may deviate from the original approaches.

For LLM+ConvoFusion, we finetune the language model to generate both speaking and listening reactions. Listening responses are represented using textual placeholders without spoken content, such as “...”, “(...listening...)”, or no text output. 
For LM-L2L Adapted, we further include chunks with speaking reactions, but disregard textual responses. For the speech-based SOLAMI, we further include chunks with listening reactions and constrain the model to not generate any speech tokens for listening reactions.

As shown in \cref{tab:baselines_alternative}, LLM+ConvoFusion achieves significant improvements on speech-related metrics, as expected due to its re-grounding on the topic of the dataset. However, motion-related metrics show only marginal gains, a pattern also observed in LM-L2L Adapted. 
Since both methods rely solely on past conversation in text, these results again underscore the importance of incorporating group motion observations to generate contextually aligned motion reactions in polyadic interactions.
SOLAMI shows degradation in MPJPE, FID, and BeatAlignDiff, with only minor improvements in speech metrics, suggesting that architectures designed for simpler settings cannot be directly extended and applied to address our task. 

Compared to all baselines and their stronger variants, \methodname remains the most competitive method, consistently ranking first or second across all speech and motion metrics, except for Diversity.

\begin{table}[t]
\caption{Impact of different motion representations. For controlled experiments on motion tokens, the pose fusion module is disabled. Diversity closer to the ground truth (117.09) indicates better performance. Root and MPJPE are reported in millimeters (mm).}
\centering
\begin{adjustbox}{width=\linewidth,center}
\begin{tabular}{cccccc}
\toprule
\makecell[c]{Pose\\Fusion} & \makecell[c]{Motion\\Representation} & Root$\downarrow$   & MPJPE$\downarrow$ & FID$\downarrow$ & Div.$\rightarrow$\\ 
\midrule
\multirow{2}[1]{*}{\cmark}&3D keypoints   &  126.9  & 150.8  &  13.82  & 123.05 \\

& transl.+rotations & \textbf{108.7} & \textbf{144.9} & \textbf{12.18} & 113.32 \\
\midrule
\multirow{2}[1]{*}{\xmark}& motion tokens  & 316.4   & 177.8 &  16.00  &  80.60 \\
 &   transl.+rotations & \textbf{124.6} & \textbf{152.5} & \textbf{13.53} & \textbf{123.19}  \\
\bottomrule
\end{tabular}
\end{adjustbox}
\label{tab:motion_repr}
\end{table}

\subsection{Motion Representation} 

In \methodname, poses are represented using global translation and 6D rotations \cite{zhou2019continuity} of body and hand joints. We further investigate the impact of alternative motion representations such as 3D keypoint positions and motion tokens.

\myheading{3D Keypoint Positions} 
While 3D keypoint positions are a commonly used pose representation \cite{mughal2024convofusion, tanke2023social}, the lack of constraints between joints introduces inconsistencies in body shapes, which results in temporal instability. As shown in \cref{tab:motion_repr}, using 3D keypoint positions (first row) results in a slightly worse performance compared to \methodname (second row).

\myheading{Motion Tokens} 
Language-model-based motion generation often relies on pre-trained motion tokenizers \cite{zhang2023generating, jiang2024solami, ng2023text2listen, ribeiro2024motiongpt, ng2024audio}, which facilitate adaptation of LLMs to the newly added motion modality.
However, the learned codebooks often struggle to generalize to unseen motions due to the limited codebook vocabularies.
This limitation reduces both performance and motion diversity, as seen in the third row of \cref{tab:motion_repr} and for SOLAMI in \cref{tab:compare_baseline_std}. In contrast, \methodname directly learns motion embeddings from global translation and 6D joint rotations, avoiding information loss caused by constrained motion vocabularies.

\begin{table*}[t]
\caption{Impact of modality ordering. Diversity closer to the ground truth (117.09) indicates better performance. Root and MPJPE are reported in millimeters (mm). Motion, social cue, and speech are denoted as \textbf{M}, \textbf{C}, and \textbf{S}, respectively.}
\centering
\begin{tabular}{lccccccccc}
\toprule
\multirow{3}[1]{*}{Modality Order} & \multicolumn{5}{c}{Motion}  & \multicolumn{3}{c}{Speech} & \multicolumn{1}{c}{State} \\ \cmidrule(lr){2-6} \cmidrule(lr){7-9} \cmidrule(lr){10-10}
                           & Root$\downarrow$   & MPJPE$\downarrow$ & \multirow{2}{*}{FID$\downarrow$} & \multirow{2}{*}{Div.$\rightarrow$} & BeatAlign & BERT  & \multirow{2}{*}{WER$\downarrow$} & \multirow{2}{*}{SIM$\uparrow$}  &  \multirow{2}{*}{AP$\uparrow$} \\ 
                           & (mm)   & (mm) &   &  & Diff.$\downarrow$ & Score$\uparrow$ & &  & \\

\midrule
\textbf{M} $\rightarrow$ \textbf{C} $\rightarrow$ \textbf{S} $\parallel$ \textbf{M} $\rightarrow$ \textbf{S}      & 135.1   & 154.9  &  14.23  & 123.94 &  0.018  & 0.502 & \textbf{1.403} & 0.638 & 0.59 \\
\textbf{S} $\rightarrow$ \textbf{M} $\rightarrow$ \textbf{C} $\parallel$ \textbf{S} $\rightarrow$ \textbf{M} (Ours)  & \textbf{108.7} & \textbf{144.9} & \textbf{12.18} & \textbf{113.32} & \textbf{0.007}  & \textbf{0.508}  & 1.436 & \textbf{0.642}& \textbf{0.67}\\
\bottomrule
\end{tabular}
\label{tab:modal_order}
\end{table*}

\subsection{Modality Order}  
For an auto-regressive model such as Llama3-8B-Instruct \cite{grattafiori2024llama}, data ordering introduces modality dependencies. We hypothesize that speech provides a stronger grounding for body movements. Therefore, the text and speech style are processed and generated before body and hand motions. To test this, we also investigate the reverse order, where body and hand motions are processed and generated before speech. As shown in \cref{tab:modal_order}, this alternative (first row) performs worse than \methodname (second row), supporting our hypothesis that conditioning motion on accompanying speech improves motion quality.

\subsection{Social Cue Embedding Length}  
In \methodname, the social cues observed from the past $\text{H}^{\text{m}}$ frames are compressed into embeddings with length of two. In \cref{tab:socaicue_length_repr}, we evaluate the effect of different embedding lengths. The results show that a length of $n=2$ achieves the best performance on motion quality, while only causing a minor reduction in Diversity. 

\begin{table}[t]
\caption{Impact of social cue embedding lengths $n$. Diversity closer to the ground truth (117.09) indicates better performance. Root and MPJPE are reported in millimeters (mm).}
\centering
\begin{tabular}{lcccc}
\toprule
 & Root$\downarrow$   & MPJPE$\downarrow$ & FID$\downarrow$ & Div.$\rightarrow$\\ 
\midrule
n = 1  & 118.8   & 152.3  &  13.56 & 123.84 \\
n = 4 & 113.7   & 149.8  &  13.04  & \textbf{120.43} \\
\midrule
n = 2 (ours) & \textbf{108.7} & \textbf{144.9} & \textbf{12.18} & 113.32 \\
\bottomrule
\end{tabular}
\label{tab:socaicue_length_repr}
\end{table}

\subsection{Interaction Role Generalization}  
To evaluate \methodname’s generalization across roles, we train it on a non-DM participant that exhibits lower activity levels. As shown in \cref{tab:nonDM}, \methodname outperforms the baselines across all metrics, demonstrating its effectiveness for participants with different roles and activity patterns.

\begin{table}[t]
\caption{Comparison of \methodname with two retrieval-based baselines on a non-DM participant.}
\centering
\setlength{\tabcolsep}{2.0pt}
\begin{adjustbox}{width=\linewidth,center}
\begin{tabular}{lccccc}
\toprule
Non-DM & MPJPE$\downarrow$ & FID$\downarrow$ & BERTScore$\uparrow$ & WER$\downarrow$ & State AP$\uparrow$ \\
\midrule
Random      & 348.4 & 24.93 & 0.513 & 1.423 & 0.50 \\
NN          & 329.6 & 21.27 & 0.526 & 1.299 & 0.50 \\
\midrule
\methodname & \textbf{288.6} & \textbf{13.54} & \textbf{0.543} & \textbf{1.251} & \textbf{0.74} \\
\bottomrule
\end{tabular}
\end{adjustbox}
\label{tab:nonDM}
\end{table}

\subsection{Long-term Generation}  
We assess long-duration generation through recursive inference over 4 and 8 rounds ($\sim$10.24s and $\sim$20.48s), where the model’s generated motion and speech are fed back as inputs to produce successive reactions. As shown in \cref{tab:multiturn}, the model preserves temporal and social coherence without motion collapse, while exhibiting gradual degradation in MPJPE, FID, and State AP over successive iterations. No catastrophic failure or physically implausible motion is observed in the visualization. We note that this evaluation is conservative, as the behaviors of other participants are kept fixed rather than fully interactive. Nevertheless, the results demonstrate the feasibility of multi-round generation and highlight a promising direction for future work.
\begin{table}[t]
\caption{Long-duration generation via recursive inference over 4 and 8 rounds ($\sim$10.24s and $\sim$20.48s).}
\centering
\setlength{\tabcolsep}{2.0pt}
\begin{adjustbox}{width=\linewidth,center}
\begin{tabular}{lccccc}
\toprule
 & MPJPE$\downarrow$ & FID$\downarrow$ & BERTScore$\uparrow$ & WER$\downarrow$ & State AP$\uparrow$ \\
\midrule
1-round (ours) & \textbf{144.9} & \textbf{12.18} & \textbf{0.508} & \textbf{1.436} & \textbf{0.67} \\
4-round      & 172.5 & 15.41 & 0.472 & 1.468 & 0.61 \\
8-round      & 179.5 & 15.60 & 0.472 & 1.487 & 0.57 \\
\bottomrule
\end{tabular}
\end{adjustbox}
\label{tab:multiturn}
\end{table}

\subsection{Speaking State and Outputs Alignment}  
We analyze whether the model learns correlations between speaking states and generated content through a shared latent space without explicit constraints. Empirically, \methodname generates an average of 5.07 words in speaking states (ground truth: 6.12) and 0.6 words on average in listening states (ground truth: 0), indicating strong alignment between predicted states and multimodal outputs. For this analysis, we use a threshold of 0.5 to distinguish between speaking and listening states.

\subsection{User Study}
In \cref{tab:user_study_question}, we list the questionnaire items used to evaluate different aspects of human perception. The turn-taking metric for user study was omitted because pilot tests showed it added cognitive load without reliable judgments. Thus, we demonstrate turn-taking in the supplementary video.

\begin{table}[t]
    \caption{User study questions for evaluating motion, speech, and overall experience.}
    \centering
    \begin{adjustbox}{width=\linewidth,center}
    \begin{tabular}{cl}
    \toprule
     Aspect & Statements  \\
    \midrule
    Motion Coherence & The motion aligns well with the speech. \\
    Motion Continuity & The movement looks continuous and real.  \\
    \midrule
    Speech Semantics & The response is relevant to the current topic of the group.   \\
    Speech Tone  &The tone of voice is natural. \\
    \midrule
    Overall &   The character reacts naturally.   \\
    \bottomrule
    \end{tabular}
    \end{adjustbox}
    \label{tab:user_study_question}
\end{table}


\end{document}